%% file: main.tex
\documentclass[10pt,conference]{IEEEtran}
\usepackage{cite}
\usepackage{amsmath,amssymb,amsfonts}
\usepackage{graphicx}
\usepackage{textcomp}
\usepackage{xcolor}
\usepackage[hyphens]{url}
\usepackage{fancyhdr}
\usepackage{hyperref}

\usepackage{array}
\usepackage{stfloats}
\usepackage{url}
\usepackage{verbatim}
\usepackage{booktabs}
\usepackage{multirow}
\usepackage{makecell}
\usepackage{caption}
\usepackage[table]{xcolor}
\usepackage{algorithm}
\usepackage{algpseudocode}  
\usepackage{subcaption} 
\usepackage{threeparttable}
\usepackage{siunitx}

\usepackage[most]{tcolorbox}
\usepackage{xcolor}

\definecolor{lightblue}{RGB}{235,245,255}
\definecolor{darkblue}{RGB}{40,90,150}
\definecolor{lightgraybox}{RGB}{248,248,248}
\definecolor{grayframe}{RGB}{120,120,120}

\newtcolorbox{insightbox}{
  colback=lightblue,
  colframe=darkblue,
  boxrule=0.5pt,
  arc=0.5mm,
  left=2pt,
  right=2pt,
  top=1pt,
  bottom=1pt,
  width=\columnwidth,
  before skip=6pt,
  after skip=6pt
}

\input{Floats/figure}
\input{Floats/table}

\input{Floats/algorithm}

\title{ELMoE-3D: Leveraging Intrinsic Elasticity of MoE for Hybrid-Bonding-Enabled Self-Speculative Decoding in On-Premises Serving}

\author{
\IEEEauthorblockN{
Yuseon Choi\textsuperscript{1,3},
Jingu Lee\textsuperscript{1},
Jungjun Oh\textsuperscript{1},
Sunjoo Whang\textsuperscript{1},
Byeongcheol Kim\textsuperscript{2},\\
Minsung Kim\textsuperscript{1},
Junhyuk Lee\textsuperscript{1},
Hoi-Jun Yoo\textsuperscript{1},
Sangjin Kim\textsuperscript{3}
}
\IEEEauthorblockA{
\textsuperscript{1}KAIST, Republic of Korea \quad
\textsuperscript{2}Samsung Electronics, Republic of Korea \quad
\textsuperscript{3}GIST, Republic of Korea
}

\IEEEauthorblockA{
\{yuseon.choi,guya102,ojj1245,sjwhang,minsung.kim,05wnsgur,hjyoo\}@kaist.ac.kr\\
bc27.kim@samsung.com \quad sangjinkim@gist.ac.kr
}
}

\begin{document}
\maketitle

\begin{abstract}

Mixture-of-Experts (MoE) models have become the dominant architecture for large-scale language models, yet on-premises serving remains fundamentally memory-bound as batching turns sparse per-token compute into dense memory activation. Memory-centric architectures (PIM, NMP) improve bandwidth but leave compute underutilized under MoE's low arithmetic intensity on high batch sizes. Speculative decoding (SD) trades idle compute for fewer target invocations, yet verification must load experts even for rejected tokens, severely limiting its benefit in MoE especially on low batch sizes. We propose ELMoE-3D, a hybrid-bonding (HB)-based HW–SW co-designed framework that unifies cache-based acceleration and speculative decoding to offer overall speedup across batch sizes. We identify two intrinsic elasticity axes of MoE—expert and bit—and jointly scale them to construct Elastic Self-Speculative Decoding (Elastic-SD), which serves as both an expert cache and a strongly aligned self-draft model accelerated by high HB bandwidth. Our LSB-augmented bit-sliced architecture exploits inherent redundancy in bit-slice representations to natively support bit-nested execution. On our 3D-stacked hardware, ELMoE-3D achieves an average 6.6× speedup and 4.4× energy efficiency gain over naïve MoE serving on xPU across batch sizes 1–16, and delivers 2.2× speedup and 1.4× energy efficiency gain over prior accelerator baselines.

\end{abstract}

\section{Introduction}

Transformer-based large language models (LLMs) have achieved remarkable success under the scaling law~\cite{scaling_law}. Increasing parameter count improves performance but proportionally increases computational cost. Mixture-of-Experts (MoE) addresses this challenge by sparsely gating feed-forward network (FFN) layers, increasing model capacity while maintaining computational efficiency~\cite{sparsely_gated}. Today, MoE architectures are widely adopted across compact edge-deployable models and server-scale deployments, making efficient MoE inference acceleration essential.

Recent MoE models around 30B parameters~\cite{deepseek_v2, qwen3, gpt_oss, glm_4_5} are attractive for privacy-sensitive on-premises applications, such as institutional chatbots and personal assistants, where a small number of concurrent users require low-latency inference without cloud access. Personal AI workstations such as NVIDIA DGX Spark and Apple Mac Studio~\cite{dgx_spark, apple_mac_studio} can host such models locally, but efficient MoE serving remains challenging. Since experts are selected independently for each token, batching and prefill collectively activate most experts. As a result, MoE computation remains sparse per token, yet its memory traffic becomes effectively dense across the batch.

To address this bandwidth bottleneck, memory-centric architectures have been explored.  Processing-in-memory (PIM)~\cite{neupims, duplex} and near-memory processing (NMP)~\cite{stratum, h2llm} alleviate raw bandwidth limitations, but high-batch routing concentrates tokens onto hot experts and exposes their limited compute throughput. 3D-IC with hybrid bonding~\cite{hb_vision, 3d_meta_lita} provides a fast memory tier with high bandwidth and throughput, reframing MoE acceleration as expert caching under limited capacity. However, independent batched requests weaken expert locality, motivating algorithmic support that increases expert reuse across decoding stages.

Speculative decoding (SD)~\cite{sd_orig} offers an algorithmic complement by drafting multiple tokens and verifying them in a single target execution. However, SD is misaligned with MoE's sparse activation: verification must load experts for all drafted tokens, including rejected ones, inflating memory traffic---especially at low batch sizes where verification-phase expert activation far exceeds autoregressive decoding (AR)~\cite{moesd, cascade, smolpu}. Prior 3D-IC-based approaches~\cite{3d_reram_sd, 3d_toksim} reduce draft cost but still suffer from high verification overhead, as models small enough to fit in the caches exhibit low accept lengths.

\figone

To address this, we propose ELMoE-3D, which unifies expert caching and self-speculative decoding by reusing the same HB-resident expert slices across AR, drafting, and verification, as illustrated in Fig.~1. Unlike independent drafting, Elastic-SD introduces no additional draft-weight footprint in the fast-memory tier. We exploit cache-aware expert routing~\cite{cache_prior, slicemoe, moe-spec} and bit-width scaling~\cite{matryoshka_quant, anyprecision, matqptq, revolver} to fit more target-aligned expert slices in HB, turning limited fast memory into a reusable tier across batch sizes.

Building on this unified design, ELMoE-3D is realized through three key HW–SW co-design components:

\begin{itemize}

    \item \textbf{Elastic Self-Speculative Decoding (Elastic-SD):} We exploit expert and bit-width elasticity to form a fully nested self-draft, preserving target routing alignment while reusing the same expert slices for AR/draft/verify.
    \item \textbf{LSB-Augmented Bit-Sliced Architecture:} We exploit the sign-extension path in bit-sliced MACs to add a half-bin offset during MSB-only execution, enabling bit-nested quantization.
    \item \textbf{Elasticity-Aware Execution Engine:} We design a phase-aware execution flow that schedules mixed-precision computation across HB and external memory tiers to efficiently serve each SD phase.
    
\end{itemize}

\section{Background}

\subsection{Mixture-of-Experts (MoE)}

As shown in Fig.~2 (left), Mixture-of-Experts (MoE) is a model architecture built on sparsely gated feed-forward network (FFN) layers, where each token selectively activates top-$k$ experts based on routing scores, scaling parameter capacity while preserving computational efficiency~\cite{sparsely_gated}. 

Prior works exploit expert locality across decoding steps, using prefetching and caching to accelerate MoE inference under similar contexts or tasks~\cite{pregated_moe, moe_infinity}. More recently, several works have shown that MoE routing has tolerance to cache-aware expert substitution or restriction, allowing memory access to be reduced without strictly following the original top-$k$ selection~\cite{cache_prior, slicemoe}.

However, such locality- and sparsity-driven optimizations become less effective under batching. Because expert selection is independent across tokens, different tokens in a batch activate different experts simultaneously. Although per-token computation remains sparse, batching raises the number of unique experts activated per step, creating a severe bandwidth bottleneck in memory-constrained on-premises systems. 

\figtwo

\subsection{Speculative Decoding (SD)}

Speculative Decoding (SD) accelerates generation with a lightweight draft model that proposes multiple candidate tokens, verified by the target in a single forward pass (Fig.~2, right). It is especially effective in memory-bound regimes, where batching turns otherwise idle compute into extra tokens per step. Recent tree-based methods~\cite{eagle2} expand a token tree by batching width-$w$ candidates over $d$ depth steps. Top-scoring nodes are selected as \textit{verify tokens} and evaluated in parallel with attention masks. With \textit{accept length} as the number of consecutively accepted tokens and one additional \textit{bonus token} from the verification path, the overall speedup can be expressed as:
\[
\text{speedup} =
\frac{(1 + \text{accepted length}) \cdot \text{latency}_{autoregressive}}
{d \cdot \text{latency}_{draft} + \text{latency}_{verify}}
\]

SD methods can be categorized into \textit{independent drafting} and \textit{self-drafting}. Independent methods use a separate draft model: EAGLE3~\cite{eagle3} employs a lightweight head distilled from the target, while SLM-SD~\cite{slm_sd_alignment} uses a smaller model from the same family. Although these methods offer low draft cost, their weaker target alignment limits accept length, and separate parameters and KV caches incur additional storage. In contrast, Self-SD constructs the draft as a subset or approximation of the target. Quantization-based Self-SD~\cite{ml_speqd, subspec, moe_speq} improves alignment using a low-precision target copy, but still duplicates weights and cannot reuse the draft representation for AR or verification caching.

\figthree

\subsection{3D-IC and Hybrid Bonding (HB)}

Hybrid Bonding (HB)-based 3D-ICs directly bond copper pads and dielectric without bumps, enabling sub-10$\mu$m pitch and substantially higher I/O density than micro-bump approaches~\cite{hb_vision, 3d_meta_tony}. As shown in Fig.~3 (left), Die-to-Wafer (D2W)~\cite{d2w} bonding integrates individually tested Known Good Dies (KGDs), improving manufacturing yield and design margin.

Fig.~3 (right) shows our HB-based xPU, where each DRAM bank is directly connected to one PE through short, wide, low-frequency links without high-frequency bank-interleaving channels~\cite{hb_r_ener}. This reduces access energy while preserving high bandwidth in total, but requires bank-wise parallelism to fully utilize the HB bandwidth. The stacked DRAM therefore serves as a capacity-limited fast-memory tier for frequently reused, low-arithmetic-intensity data, while off-package LPDDR5 supplies residual capacity~\cite{3dvcache}.



\section{Motivation}

\subsection{Low Arithmetic Intensity in MoE Serving}

\figfour

In LLM inference, arithmetic intensity of dense layers scales linearly with batch size due to weight sharing across tokens. In contrast, MoE expert layers process only a subset of tokens, and effective batch per expert scales with $bs \times \lambda$ ($\lambda = N_{\text{active}} / N_{\text{total}}$), rather than $bs$. Consequently, even at large batch sizes, expert layers remain memory-bound while dominating inference latency due to their large weight size, as shown in Fig.~4(a). Prior PIM/NMP architectures provide high internal bandwidth but limited compute throughput. As batch size grows, expert activation exceeds their internal processing capability, forcing offload through the narrow external interconnect~\cite{neupims}. This motivates a unified HW--SW approach that increases effective arithmetic intensity.


\subsection{Misalignment between MoE and SD}

Speculative Decoding (SD) achieves speedup by drafting multiple tokens and verifying them in a single forward pass. However, in MoE models, verification introduces a fundamental inefficiency: experts are activated for all drafted tokens, including those that are ultimately rejected. As shown in Fig.~4(b) (left), this leads to a substantial increase in the number of activated experts compared to autoregressive decoding.

This mismatch leads to a batch-size-dependent efficiency trend even after accounting for SD acceleration (Fig.~4(b), right). At low batch sizes, SD can still be less efficient than AR because verification activates excessive experts. As batch size increases, SD becomes favorable once AR memory access becomes comparable to verification overhead and can be amortized across accepted draft tokens.

Although 3D-IC accelerates draft execution, independent draft models consume limited fast-memory capacity without improving target expert caching. Quantization-based self-drafting improves target alignment but still requires a separate low-precision representation. Rather than eliminating the expert accesses required for exact verification, we aim to reduce their memory cost by sharing one HB-resident representation across caching and self-drafting. This motivates a capacity-efficient self-draft constructed directly from cached target experts.



\figfive

\subsection{Intrinsic Elasticity of MoE}

To build a draft that fits in HB memory while remaining aligned with the target MoE, we exploit elasticities.


\textbf{Expert Elasticity.}
MoE routing has flexibility because high-score experts near the top-$k$ boundary can often serve as near-substitutes~\cite{cache_prior, lightmoe}. As shown in Fig.~5(a) left, ELMoE-3D exploits this through \textit{expert throttling}, forming a hot expert subset across requests at each draft phase and restricting draft top-$k$ routing to this subset during the draft depth. Since tokens still follow the target router's score ordering within the subset, missing top-$k$ experts are replaced by high-scoring resident experts rather than unrelated experts. This preserves much of the target routing behavior with fewer activated experts.

\textbf{Bit-width Elasticity.}
Bit-width scaling such as quantization preserves weight vector direction, which is advantageous for draft--target alignment. However, conventional integer quantization uses different scaling factors per bit-width, requiring duplicated storage for multiple precisions. As depicted in Fig.~5(a) right, bit-nested quantization resolves this by designing the most-significant-bit (MSB) slice of an INT8 weight to be itself a valid low-bit weight, enabling precision switching via simple bit truncation without memory duplication.

\textbf{Two-Axis Exploration.}
As shown in Fig.~5(b), the two axes are complementary under limited HB capacity. Bit-width elasticity reduces each expert's footprint, allowing expert throttling to keep more hot experts resident and better preserve the target routing distribution. The red region marks HB-cacheable configurations, and the yellow star selects the point that minimizes draft cost while maintaining high acceptance ratio. This joint scaling forms a fully nested self-draft that fits in HB while preserving routing alignment.


\section{Proposed ELMoE-3D Architecture}

\figsix

\subsection{Design Overview}

ELMoE-3D is a hardware--software co-designed framework for on-premises MoE serving on hybrid-bonding-based xPU. It targets batch sizes of 1 to 16~\cite{h2llm, onprem16} and $\sim$30B-scale MoE models under INT8 quantization, a practical inference setting for recent on-premises accelerators~\cite{dgx_spark, hexagon, amd_ryzen, mobilint}.

\textbf{Overall Architecture.}
Fig.~6 (left) illustrates the system, which comprises two heterogeneous memory tiers: 4 or 8GB LPDDR-based hybrid-bonded (HB) memory and 64GB LPDDR external (EXT) memory. This hierarchy is designed to keep frequently reused expert slices in HB while leaving residual capacity to EXT. Each PE is paired with an HB bank for high-bandwidth low-energy access, while external memory connects via a lower-bandwidth LPDDR PHY. PEs are organized into groups with aggregation links, and a streamlined interconnect circulates across PE groups.


\textbf{HB-based Microarchitecture Design.}
In Fig.~6 (center), each PE fetches data from its HB bank via an HB controller and from external memory via the streamline. The PE includes 32KB WMEM and 32KB IOMEM, driving a $16\times32\times32$ tensor PE array. A bit-sliced unit PE supports multi-precision weights (4--8 bit) with 8-bit activations, avoiding area-heavy precision-specific cores while maintaining high utilization. A vector processing unit (VPU) handles element-wise operations and non-GEMM workloads. Partial sums are accumulated through a 32-way adder tree and immediately scaled, fusing dequantization into the accumulation path.

\textbf{Logic Die Datapath and Control.}
Fig.~6 (right) shows the datapath, where a memory controller orchestrates data flow from both tiers. HB data is directly routed to local PE buffers, while EXT data circulates across PE groups through the streamline. The streamline provides high-bandwidth weight delivery, whereas the aggregation link supports flexible communication (e.g., all-to-all, all-reduce) with buffering, bypass, and accumulation support. They strengthen inter-PE connectivity to preserve bank-wise parallelism during HB-resident expert execution.

\figseven

\subsection{HB-enabled Elastic Self-SD}

\textbf{Elastic-SD.}
Elastic-SD leverages the remaining HB capacity to host a fully nested draft model. Each draft phase uses an HB-resident hot expert subset (expert pool) represented by 4-bit MSB slices, and applies the same MSB-only execution to static weights and shared experts in the draft path. Drafting restricts top-$k$ routing across all draft iterations (e.g., 10 steps) to this subset, eliminating external memory accesses and significantly reducing draft latency. As depicted in Fig.~7, when an originally preferred expert (e.g., Expert~6) is unavailable, the highest-scoring candidate within the subset (e.g., Expert~3) is selected instead. Since experts with similar routing scores tend to make similar contributions, this score-preserving substitution maintains draft quality.

\textbf{Expert Throttling.}
We term this HB-capacity-aware subset selection \emph{expert throttling}. The HB capacity remaining after KV-cache preallocation determines the resident expert count, and throttling selects the largest hot subset that fits. Hotness is estimated from one-hot routing decisions accumulated during the preceding draft phase. Because draft tokens from the same request exhibit routing locality, the selected subset is reused throughout the draft window.


\algone

\textbf{AR/SD Mode Selection.}
ELMoE-3D selects AR or SD using a model-specific batch threshold calibrated offline. GPT-OSS-20B uses AR throughout the evaluated batch range, while the remaining models switch to SD at batch size $\tau=3$, requiring no runtime latency or acceptance prediction.

\textbf{Expert Pool Update.}
The expert pool must be refreshed before the next draft phase to track new content or requests. A separate update is inefficient because HB and EXT lack a direct link, preventing the required expert access from being pipelined with verification. Thus, the next subset is determined in the previous draft phase using accumulated one-hot routing vectors, so its MSB slices can be written into HB during verification by reusing streamed expert data. Context similarity hides most updates, while new-context arrivals may reduce overlap but occur infrequently. Algorithm 1 summarizes the execution: the system selects AR or SD based on the calibrated threshold; SD generates candidates using HB-resident MSB slices and throttled routing, verifies them with the original INT8 target model, and refreshes the expert pool during verification for the next draft phase.

\figeight

\subsection{LSB-Augmented Bit-Sliced Architecture for Bit-Nested Quantization}

Elastic-SD requires bit-nested quantization because the HB-resident 4-bit draft weights must be reused as the MSB prefix of the 8-bit target weights. A separate 4-bit quantized draft copy would consume HB capacity, while naive truncation of uniform INT8 weights yields a severe clipping error. Prior works address flexible precision through quantization-aware training (QAT)~\cite{matryoshka_quant}, non-uniform quantization (NUQ)~\cite{anyprecision, anybcq}, or calibration-based PTQ~\cite{matqptq}. However, these approaches require training/calibration or introduce metadata and arithmetic overhead. We therefore target metadata-free bit nesting in a hardware-friendly, plug-and-play manner.

\textbf{Revisiting Bit-Slice Redundancy.}
In a conventional bit-slice architecture~\cite{bitfusion, sibia, panacea, bada, sevedo}, a signed 8-bit weight is split into two slices fed into the same signed$\times$signed arithmetic unit. The least-significant-bit (LSB) slice, originally unsigned, requires an auxiliary 1-bit zero-extension to become signed, while the MSB slice undergoes sign extension that carries no new information.

As shown in Fig.~8, directly using the MSB slice as a 4-bit weight truncates the lower 4 bits, causing a one-sided approximation error. Conventional rounding reduces this error by carrying into the MSB slice, but the residual may exceed the range of the original 4-bit LSB slice. Exact 8-bit reconstruction therefore requires carry metadata or a wider residual slice, as in prior designs~\cite{sibia, nestquant}. Such extra storage increases the expert footprint, reducing HB-cacheable experts and increasing external-memory access latency.

\textbf{LSB Augmentation.}
Instead of sign-extending the MSB slice, we left-align the data and hardwire a \texttt{1} at the LSB position of the MSB slice during MSB-only operation (draft phase) and a \texttt{0} when combining with the full 8 bits (AR/verify phase). This half-unit offset redefines the 4-bit reconstruction grid using zero-centered points on the original 8-bit scaling grid. As a result, 4-bit data is represented as a rounding approximation rather than a clipping approximation, preserving high accuracy. Thus, LSB augmentation provides the error behavior of rounding through microarchitectural compensation, while requiring no additional memory footprint and only a small multiplexer on the existing bit-slice datapath.


\section{ELMoE-3D Execution Engine}

ELMoE-3D orchestrates data placement and computation across heterogeneous memory tiers for efficient execution. 

\fignine

\subsection{Phase-wise Parallelism Exploitation}

\textbf{Coupled vs. Decoupled Memory Characteristics.}
As depicted in Fig.~9(a), HB memory connects each bank one-to-one with a PE via a dedicated datapath, making bandwidth and throughput \emph{coupled}. Engaging more PEs with unique workloads proportionally increases usable bandwidth, while idle PEs leave their corresponding bandwidth unused. External memory streams data through a single shared interconnect, making bandwidth and throughput \emph{decoupled}. Fig.~9(b) illustrates this on the system-level roofline. At partial PE utilization, HB loses both bandwidth and throughput, whereas EXT mainly loses throughput. This motivates phase-wise parallelism matched to each tier.

\figten

\textbf{Phase-wise Data--Tensor Parallelism.}
The distinct characteristics of HB and EXT memory necessitate different parallelization strategies. In HB, each bank is directly connected to one PE, so total bandwidth is realized only when different PEs consume different weight slices. HB-resident data therefore favors Tensor Parallelism (TP), where weights are partitioned across PEs and activations are replicated. In contrast, EXT provides a fixed bandwidth through a shared interconnect. When the arithmetic intensity of a streamed weight slice is high, a single PE cannot fully utilize its compute opportunity alone; the same fetched data should instead be reused across multiple PEs. EXT-resident data therefore favors Data Parallelism (DP), where weights are shared across PEs and activations are partitioned across tokens.

This asymmetry is critical for bit-sliced execution. EXT-resident LSB slices benefit from DP because streamed weights are reused across tokens, whereas HB-resident MSB slices benefit from TP because bank--PE locality exposes HB bandwidth. Forcing either slice into the other's layout would waste EXT reuse or require additional HB data paths. Accordingly, ELMoE-3D processes HB-resident MSB slices with TP and EXT-resident LSB slices and uncached experts with DP.


\subsection{Activation Communication Strategy}

\textbf{MSB--LSB Weight Asynchronous Processing.}
ELMoE-3D processes HB-resident MSB slices with TP and EXT-resident LSB slices with DP, avoiding full-weight reconstruction. As shown in Fig.~10(a), remapping MSB slices to the EXT-side DP layout, which is used under highly batched verification, would require a dedicated high-bandwidth path comparable to the 128GB/s streamline. Instead, ELMoE-3D exchanges only partial sums (PSUMs), whose traffic fits within the existing 32GB/s aggregation link even under worst-case batching.

Fig.~10(b) shows the resulting fused expert pipeline. To avoid PSUM footprint explosion, the up, gate, and down projections are fused~\cite{stratum}, so synchronization is needed only at projection boundaries and the final reduction. MSB and LSB slices are fetched and computed asynchronously, while PSUM communications are hidden under the slower EXT LSB fetch even under skewed expert activation, keeping communication off the critical path. 

\tabone

\figeleven

\tabtwo

\tabthree

\section{Implementation}

Table~1 summarizes the ELMoE-3D system specification.

\textbf{Logic Die.}
The logic die is synthesized using the ASAP7 7nm FinFET PDK~\cite{asap7} at 0.7 V and 1.0 GHz. Each PE contains a $16{\times}32{\times}32$ bit-sliced MAC array composed of IA8$\times$W5 integer unit PEs. We evaluate two configurations that scale PE count with HB DRAM capacity, pairing 16 PEs with 4GB and 32 PEs with 8GB. The 8GB configuration represents a scaled-system design point rather than an iso-area capacity expansion. Its compute and memory resources scale together to preserve the same architectural organization as the 4GB design. The aggregation link~\cite{aggr_link} (32GB/s) supports bypass, all-to-all, and accumulation for activation and PSUM transfer, while the streamline link (128GB/s) delivers external weight data directly into each PE's WMEM~\cite{lpu}.

\textbf{Memory.}
Each HB bank provides 51.2GB/s through 1024 pins at 0.4GHz, limited by the 2.5 ns $t_{\mathrm{CCD,short}}$ constraint~\cite{hbnpu}. The HB read energy of 0.43 pJ/bit is estimated by applying voltage-rail-aware scaling to the reported 0.88 pJ/bit LPDDR4-based HB measurement~\cite{hb_r_ener}, using the LPDDR5 low-voltage operating conditions (VDD2L=0.9V, VDDQ=0.5V)~\cite{jedec_lpddr5}. External memory uses LPDDR5-6400~\cite{micron_lpddr5} with 8 channels (x16), providing 102.4GB/s and 64GB capacity at 3.88 pJ/bit~\cite{lpddr5_pim}.

\textbf{Area Breakdown.}
Fig.~11 shows the area breakdown of the logic die in the 4 GB configuration. PEs dominate the design, occupying 93\% of the total area. Within each PE, the tensor compute logic primarily consists of low-bit integer MAC and dequantization units, enabling high area efficiency. Bit-sliced support logic (shifters, multiplexers) introduces negligible overhead, indicating that bit-nested execution can be supported with minimal additional cost.

\section{Evaluation}

\textbf{Models and Datasets.}
Table~2 lists the four target models. Qwen3-30B-A3B~\cite{qwen3}, GLM-Flash~\cite{glm_4_5}, DeepSeek-V2-Lite~\cite{deepseek_v2}, and GPT-OSS-20B~\cite{gpt_oss} are all $\sim$30B-class MoE models suitable for on-premises deployment, with activated parameters ranging from 2.4B to 3.6B. GPT-OSS-20B provides MXFP4-quantized expert weights, so both Quant-SD and Elastic-SD apply bit-width scaling only to dense layers, with Elastic-SD using expert throttling alone.
Datasets include MT-Bench~\cite{mt_bench} (multi-turn dialogue), GSM8K~\cite{gsm8k} (math reasoning), Alpaca~\cite{alpaca} (instruction following), and HumanEval~\cite{humaneval} (code generation). We use 1k sequence length for baseline.

\textbf{Inference Framework.}
All models use G32 INT8 symmetric quantization with FP16 scaling factors as the baseline precision. Speculative decoding schemes use a tree-based framework~\cite{eagle1} with depth 10, width 8, and 64 verify tokens per request. All evaluations use greedy decoding. Draft candidates are verified using the original INT8 target routing and weights, preserving the target's greedy output sequence. \textit{EAGLE-3}~\cite{eagle3} uses an independent lightweight draft head. \textit{SLM-SD}~\cite{slm_sd_alignment} uses Qwen2.5-1.5B-Instruct~\cite{qwen25} as a draft. \textit{Quant-SD}~\cite{subspec, moe_speq} uses a separately quantized 4-bit copy of the target. For these baselines, HB caching is applied to full-bit experts under an LRU policy~\cite{lru_approx}. Elastic-SD (Ours) caches MSB 4b expert slices and tracks verification hits at slice-pool granularity for each verify phase.

\tabfour
\tabfive

\textbf{Accelerator Baselines.}
Table~3 summarizes the baseline architectures for memory-bound MoE serving. All baselines share the same xPU compute capability and LPDDR5-based external memory for comparing memory-centric designs. \textit{xPU-PIM} adds LPDDR5-PIM near-bank compute units~\cite{lpddr5_pim}, providing high bank-internal bandwidth but limited throughput. \textit{xPU-LogicPIM} uses a Duplex-like near-memory logic die~\cite{duplex}, improving throughput with lower effective bandwidth than \textit{xPU-PIM}. \textit{xPU-NMP} follows an H2LLM-like hybrid-bonded NMP design~\cite{h2llm}, providing high bandwidth on selected channels. \textit{HB-xPU} adds a fast-memory tier with limited capacity, and ELMoE-3D further adds bit-sliced MACs and Elastic-SD for cache--draft unification. HB-aware TP/DP co-execution addresses bit-nested experts' MSB/LSB dependency, which independent-draft baselines lack. We exclude 3D-TokSim~\cite{3d_toksim} because its dense-LLM CIM dataflow relies on weight reuse unsuitable for sparse MoE activation.

Latency is evaluated using a Duplex-based trace-driven, cycle-accurate simulator~\cite{duplex}, with Ramulator 2.0~\cite{ramulator} configured for HB DRAM timing and MoE/SD execution traces. While the one-to-one bank-to-PE organization eliminates inter-PE bank contention, HB's high bandwidth makes it more sensitive to row-miss penalties. The 16KB row buffers mitigate this sensitivity through contiguous expert accesses, limiting bandwidth degradation to approximately 3\% relative to the roofline. External memory achieves within 1\% of peak bandwidth.

\subsection{SD Scheme Comparison on ELMoE-3D}

Fig.~12 (top) compares how each SD scheme occupies the HB/EXT hierarchy. Static weights and KV cache are placed in HB first, and the remaining capacity is used for draft parameters or expert caching. EAGLE-3 and SLM-SD consume HB space for separate draft weights and draft KV cache. Quant-SD stores independent 4-bit draft and 8-bit target weights, duplicating the expert footprint across both tiers. Elastic-SD reuses the target's MSB 4-bit slices as the draft, leaving more HB capacity for expert caching.

\figtwelve

\figthirteen

\figfourteen

Fig.~12 (bottom) shows how this storage difference translates into latency and speedup on MT-Bench~\cite{mt_bench}, where requests are randomly sampled from eight subjects and served under a multi-turn continuous batching scenario. EAGLE-3 and SLM-SD have low draft latency but limited accept length, while Quant-SD improves acceptance at high draft/verify cost due to duplicated draft representation. Elastic-SD avoids extra draft storage, keeping draft latency low and verification cost lower through larger effective expert-cache capacity.

Table~4 shows that Elastic-SD improves both AR and SD modes and scales with HB capacity across models and batch sizes. The key enabler is that every expert slice contributes for both AR/verification cache coverage and draft alignment. Increasing HB capacity therefore improves both target-cache coverage and draft alignment. Compared with EAGLE3 in SD mode, Elastic-SD improves $\Phi_{\text{SD}}$ by 1.30$\times$ at 4GB HB and 1.61$\times$ at 8GB HB across batch sizes of 4, 8, and 16. Overall, Elastic-SD achieves 3.4$\times$ and 7.0$\times$ average speedup over xPU-only serving at 4GB and 8GB HB. AR is selected when SD verification overhead is not sufficiently amortized, especially at low batch sizes. For GPT-OSS-20B, native low-bit experts allow many more experts to fit in the same HB capacity, extending the regime where HB-cached AR remains favorable. Table 4 separately characterizes draft quality through acceptance ratios, while Fig. 12 quantifies draft and verification costs on the proposed hardware. Together, they distinguish the drafter's alignment quality from the architectural benefits of shared expert caching.

\figfifteen
\figsixteen

Fig.~13 shows how the benefit of bit-sliced caching depends on cache capacity and batch size. Under limited HB capacity, bit-sliced caching increases the number of resident experts and reduces verification latency despite using lower-bit slices. At small batch sizes, cache pressure is low, so full-bit caching can occasionally be competitive. However, as batch size grows, the active expert set expands and cache capacity becomes the dominant bottleneck, allowing bit-sliced caching to provide lower verification latency by covering more hot experts.

Table~5 further evaluates draft alignment robustness across models, datasets, token types, and positions. Elastic-SD consistently outperforms EAGLE3 across mixed model–dataset settings, showing that the HB-resident expert pool generalizes beyond a single workload. EAGLE3 degrades on frequent/rare content-bearing tokens and mid/late draft positions, indicating sensitivity to limited model capacity and its single-layer KV representation. Elastic-SD remains stable by preserving broader target routing through resident expert slices.

\subsection{Comparison with Accelerator Baselines}

\textbf{Speedup.}
Fig.~14 reports speedup over the xPU-only baselines. Dashed lines with triangle markers denote AR mode, and solid lines denote SD mode. Prior memory-centric accelerators show advantages primarily in AR mode at small batch sizes. xPU-PIM performs well for Qwen3 and GLM, while xPU-NMP is strongest for DS2 and GPT-OSS. However, their SD-mode gains are limited, as speculative verification increases compute demand beyond their internal capability. Our design consistently achieves the highest speedup. Elastic-SD improves cache hit rates in both AR and verification, and SD-mode speedup continues to scale with batch size without compute bottleneck. Averaged across all models and datasets at 8GB, our system achieves 6.6$\times$ mean speedup over the xPU baseline (up to 12.0$\times$), and 2.2$\times$ over the strongest prior baseline (up to 9.9$\times$).

\figseventeen

\textbf{Energy Consumption.}
Fig.~15 reports per-token energy normalized to Ours, decomposed into computation, external memory access, HB memory access, communication, and static power. Blue markers indicate SD-selected configurations. As batch size grows, PIM/LogicPIM/NMP increasingly rely on xPU offload, incurring high external-DRAM energy. ELMoE-3D instead maximizes 9× more efficient HB access through Elastic-SD's AR/verify weight reuse, while lower latency reduces static energy. Overall, Ours reduces energy per token by 4.4$\times$ on average (up to 7.2$\times$) compared to the xPU baseline, and 1.4$\times$ (up to 4.9$\times$) compared to the best-performing prior accelerator.

\subsection{Ablation Studies}

Fig.~16 decomposes the contribution of ELMoE-3D across algorithm, microarchitecture, and execution support. HB-xPU alone achieves 3.11$\times$ speedup by exploiting expert locality through the HB fast-memory tier. However, directly adding a baseline draft (EAGLE3) underperforms caching, because the draft quality is insufficient to offset the reduced expert-cache capacity. Expert-only Elastic-SD lacks sufficient resident high-score experts, while bit-only Elastic-SD still pays external-memory draft cost. Combining both axes forms an HB-resident draft with higher alignment and larger expert coverage.

The remaining ablations demonstrate the need for the hardware mechanisms. Without LSB augmentation, rounded bit-nested execution requires 1-bit metadata per 4-bit LSB slice, increasing memory footprint and verification traffic. Without HB-aware TP/DP parallelism, EXT-fetched slices are restricted to single-PE TP, reducing large-batch SD performance by 29\%. Without PSUM synchronization, weight-level MSB/LSB coherence overloads the 32GB/s aggregation path, adding 10\% latency. Thus, Elastic-SD creates the cache--draft reuse opportunity, and the hardware support converts it into end-to-end speedup.

\subsection{Elastic-SD Analysis}

\figeighteen

\textbf{a. Expert Locality.}
Fig.~17 analyzes expert locality across batch size, request type, model, and benchmark. As batching expands the inter-request expert working set, AR caching alone cannot sustain low miss rates (Fig.~17a). SD, however, activates fewer unique experts for the same token count because draft tokens share context (Fig.~17b). Locality also varies more across models than benchmarks (Fig.~17c,d), indicating that expert reuse is mainly model-dependent. This motivates one HB-resident expert pool reused across draft depths.

\textbf{b. Expert-Axis Elasticity.}
Fig.~18 analyzes why expert throttling preserves draft alignment. By storing MSB 4-bit slices, ELMoE-3D caches more experts than full 8-bit caching under the same capacity (Fig.~18a). The larger resident pool keeps the routing-score sum close to original routing and improves top-k expert overlap, meaning that constrained routing still finds target-relevant substitutes (Fig.~18b,c). Hotness-aware selection improves acceptance by approximately 20–30 percentage points over random expert placement across batch sizes (Fig.~18d). Notably, this locality persists in Qwen3-30B-A3B despite its native load-balancing objective~\cite{qwen3}, demonstrating that hotness-aware selection remains effective under load-balanced routing.

\textbf{c. Bit-Axis Elasticity.}
Fig.~19 compares bit-axis design choices. Quant-SD provides a strong-alignment baseline, but it is not bit-nested and requires a separate draft copy. Truncation is metadata-free, but its severe SNR loss collapses acceptance (Fig.~19a (left)). Rounding and LSB-Aug-SD avoid most of this SNR loss, but rounding requires a 12\% memory-footprint overhead (Fig.~19a (mid)). LSB-Aug-SD preserves high acceptance without footprint overhead by exploiting bit-slice redundancy with only a small multiplexer and one control signal (Fig.~19a (right)). Across additional models and mixed datasets, LSB augmentation consistently improves draft acceptance over naive MSB truncation (Fig.~19b).

\fignineteen

\textbf{d. Expert Pool Update in Realistic Batching Scenarios}
Table~6 evaluates expert-pool refresh overhead under MT-Bench continuous batching with three request-arrival cases: \textit{decode-only}, \textit{multi-turn}, and \textit{new query}. In \textit{decode-only} iterations, no new request arrives, so context continuity hides most update traffic under verification, achieving 87\% and 99\% overlap at batch sizes 4 and 16. In \textit{multi-turn}, partial context similarity is preserved, whereas \textit{new query} introduces a less similar context; both may update more experts and temporarily reduce the arrived request's acceptance ratio. However, such arrival-boundary iterations comprise only 1.37--1.58\% of all iterations, making their amortized impact negligible. 

Table~7 further stresses request turnover by limiting decode lengths and immediately replacing completed requests from a saturated queue under mixed workloads. Shorter generations increase refresh frequency and its overhead, but acceptance degradation is confined to newly arriving requests, limiting its impact on overall acceptance. Moreover, most pool updates overlap with verification, keeping the amortized latency overhead below 0.23\%.

\section{Discussion}

From a system perspective, the primary limitation is resource contention in HB capacity. KV caches are reused at every layer and must be prioritized in HB. As batch size and sequence length increase, the growing KV footprint reduces expert-cache capacity and may lower hit rates. Table~8 confirms that from 1k to 4k sequence length, Elastic-SD's cached experts decrease and its acceptance ratio drops, showing that long contexts reduce its relative benefit despite remaining faster than EAGLE-3.

Extending Table~8 to 8k and 16k reveals a shift in memory allocation priorities. At 16k, full-KV residency exhausts HB capacity, preventing expert caching and rendering Elastic-SD infeasible. Yet not all historical KV states are equally valuable for drafting, motivating selective draft KV retention to prioritize expert caching. We demonstrate this with training-free sparse attention inspired by StreamingLLM~\cite{streamingllm}, combining sink, recent, and strided intermediate tokens while retaining full attention for exact verification. With a 4,096-token draft KV budget, Elastic-SD restores 41 cached experts at 16k, achieving 1.87$\times$ speedup over xPU and 1.72$\times$ over EAGLE-3.

\tabsix
\tabseven

Elastic-SD applies to general two-tier memory systems, but HB is particularly well matched to its access pattern, offering high local bandwidth and low read energy through dense vertical I/O and short bank-to-PE movement despite limited fast capacity. However, HB's one-to-one bank-to-PE coupling limits execution flexibility through PE-dependent bandwidth utilization. Our TP/DP co-execution addresses this constraint by exploiting bank-wise parallelism for HB-resident MSB slices while preserving data reuse for EXT-resident LSB slices.
\tabeight

The high bandwidth of HB raises thermal concerns in vertically stacked integration. Table~9 reports phase-wise power against an LPDDR5-based xPU baseline. HB-intensive drafting raises instantaneous power but occupies only 22\% of execution time, while verification shifts demand toward external memory, avoiding simultaneous saturation. Averaged over a decoding cycle, ELMoE-3D draws the same power as the baseline while delivering 4.8$\times$ the throughput, as HB displaces energy-intensive external-memory accesses---a 4.7$\times$ reduction in energy per token. The one-to-one bank-to-PE organization further distributes activity across the logic die. Together, these temporal and spatial characteristics mitigate thermal concerns despite higher drafting power.

\tabnine

\section{Conclusion}

We propose ELMoE-3D, a hybrid-bonding-based HW--SW framework that unifies caching and speculative decoding for memory-bound MoE serving. Exploiting expert- and bit-width-axis elasticity, Elastic-SD serves as both a cache and a self-draft within limited HB capacity. A bit-sliced datapath provides native multi-precision support with negligible hardware overhead, while the execution engine exploits heterogeneous memory tiers for efficient resource utilization.

Across diverse scenarios, ELMoE-3D with 8GB configuration achieves 6.6$\times$ speedup and 4.4$\times$ energy gain over the xPU baseline, and 2.2$\times$ speedup and 1.4$\times$ energy gain over the state-of-the-art memory-centric architecture. These results show that MoE's intrinsic elasticity enables efficient on-premises serving under limited HB capacity.


\bibliographystyle{IEEEtranS}
\bibliography{refs}

\end{document}

%% file: Floats/figure.tex
\newcommand{\figone}{
\begin{figure}[t]
    \centering
    \includegraphics[width=\linewidth]{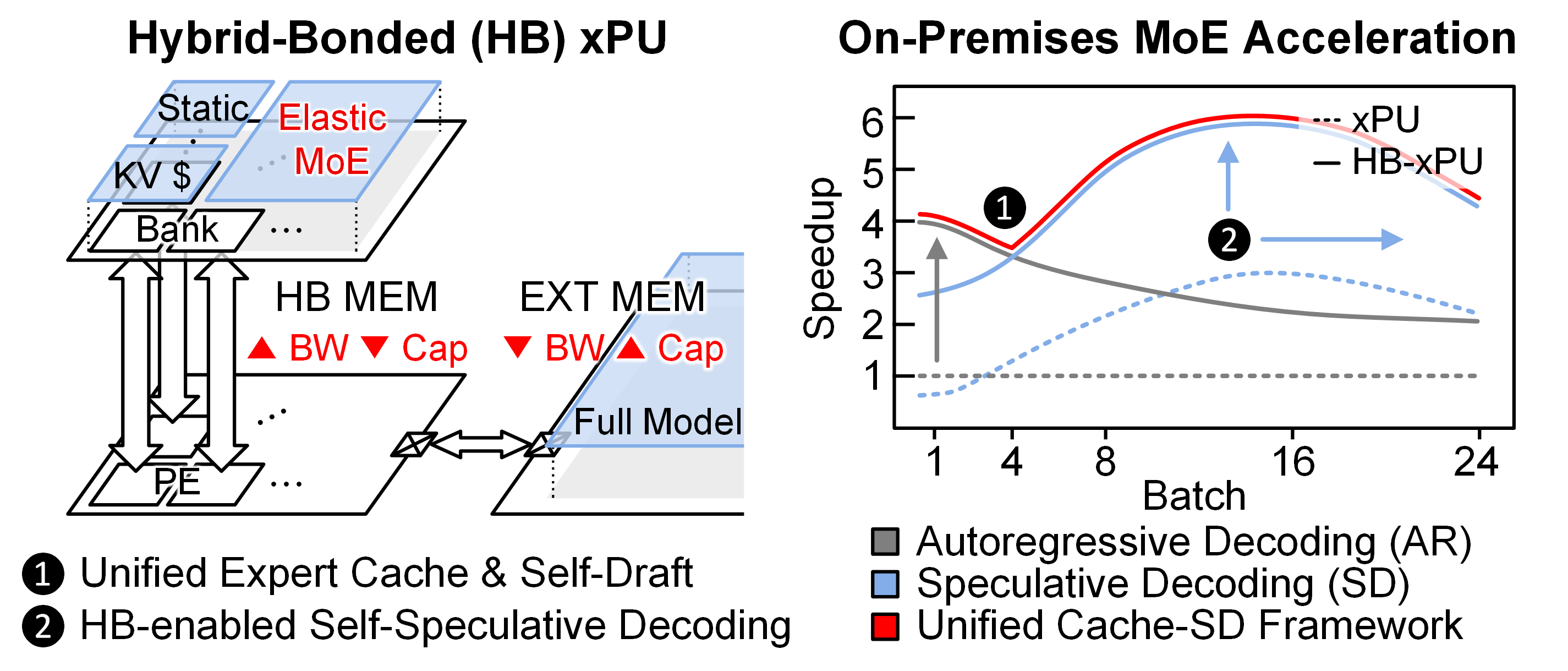}
    \caption{Overview of ELMoE-3D.}
    \label{fig_1}
    \vspace{-0.5cm}
\end{figure}
}

\newcommand{\figtwo}{
\begin{figure}[t]
    \centering
    \includegraphics[width=\linewidth]{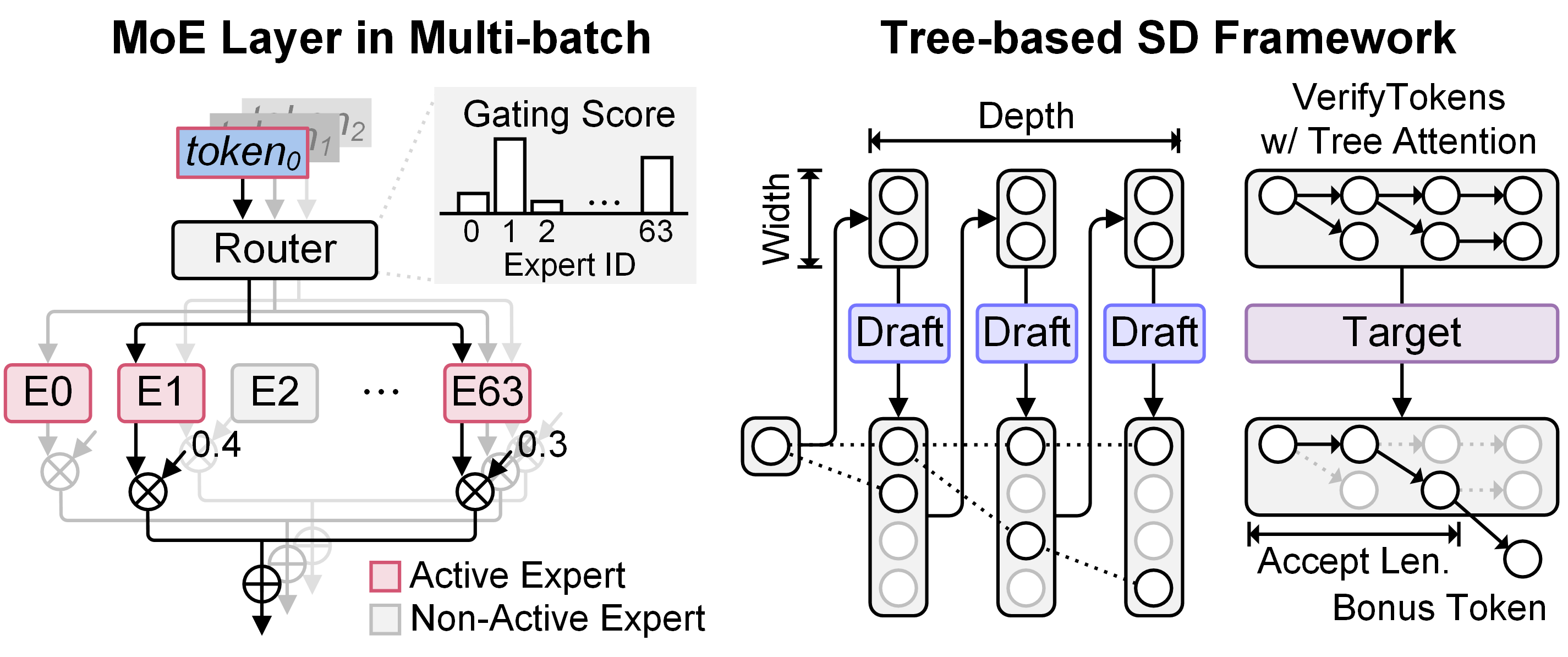}
    \caption{Mixture-of-Experts and Speculative Decoding.}
    \label{fig_2}
    \vspace{-0.5cm}
\end{figure}
}

\newcommand{\figthree}{
\begin{figure}[t]
    \centering
    \includegraphics[width=\linewidth]{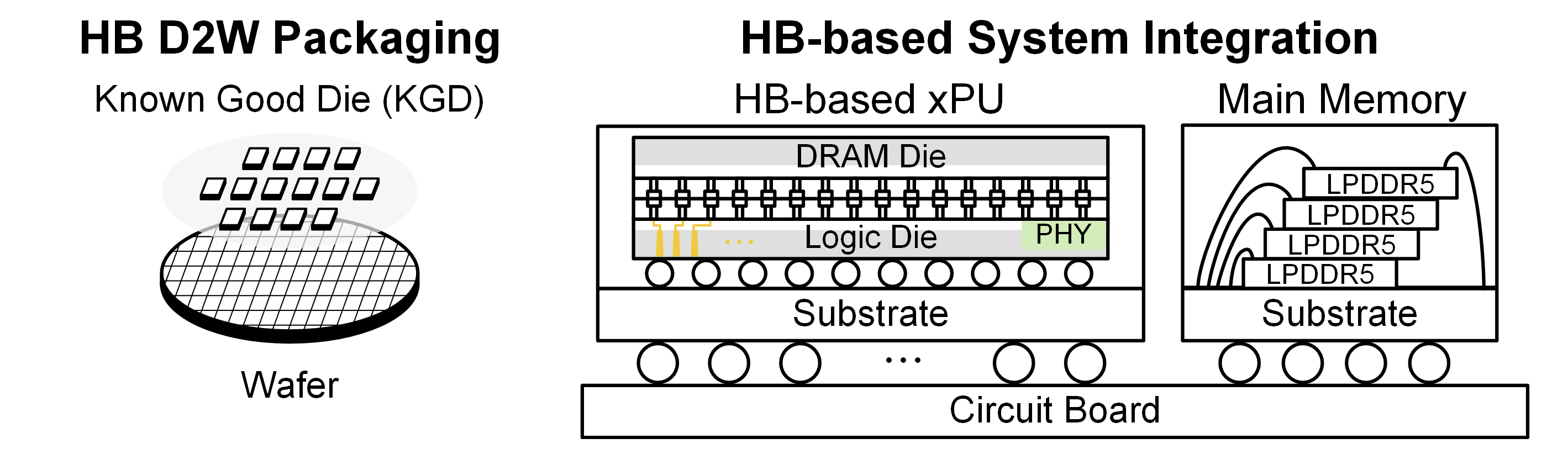}
    \caption{Die-to-Wafer hybrid bonding (HB) process and HB-based system integration.}
    \label{fig_3}
    \vspace{-0.5cm}
\end{figure}
}

\newcommand{\figfour}{
\begin{figure}[t]
    \centering
    \begin{subfigure}{\linewidth}
        \centering
        \includegraphics[width=\linewidth]{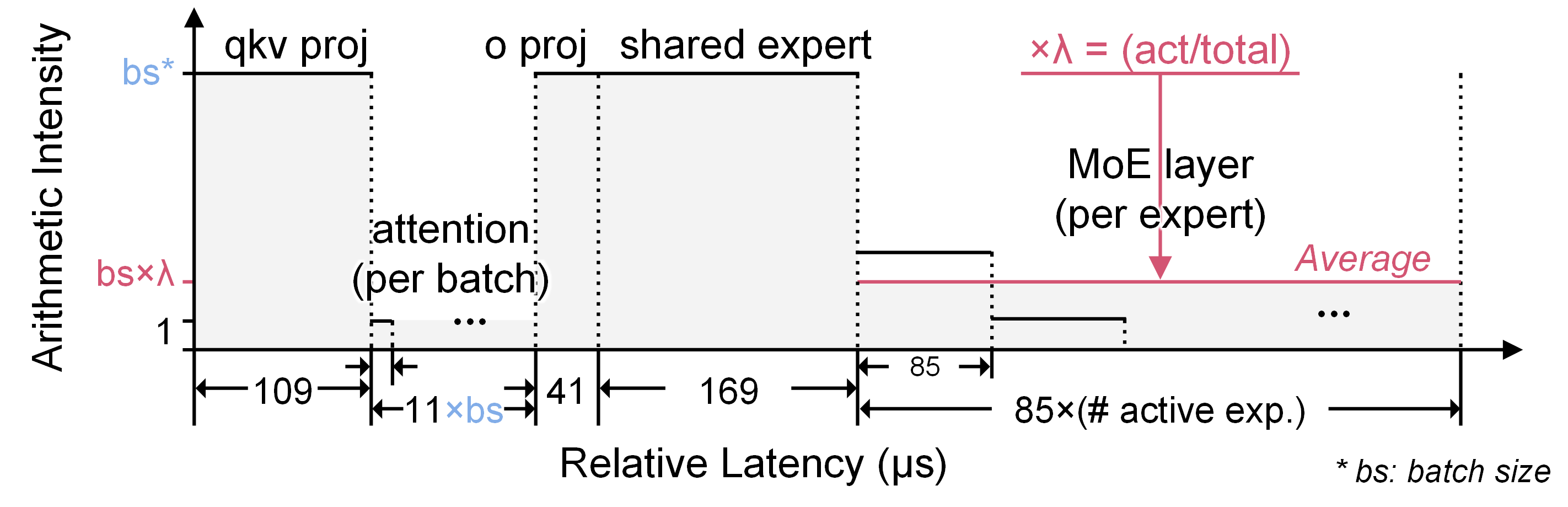}
        \caption{}
        \label{fig_4_a}
    \end{subfigure}
    \begin{subfigure}{\linewidth}
        \centering
        \includegraphics[width=\linewidth]{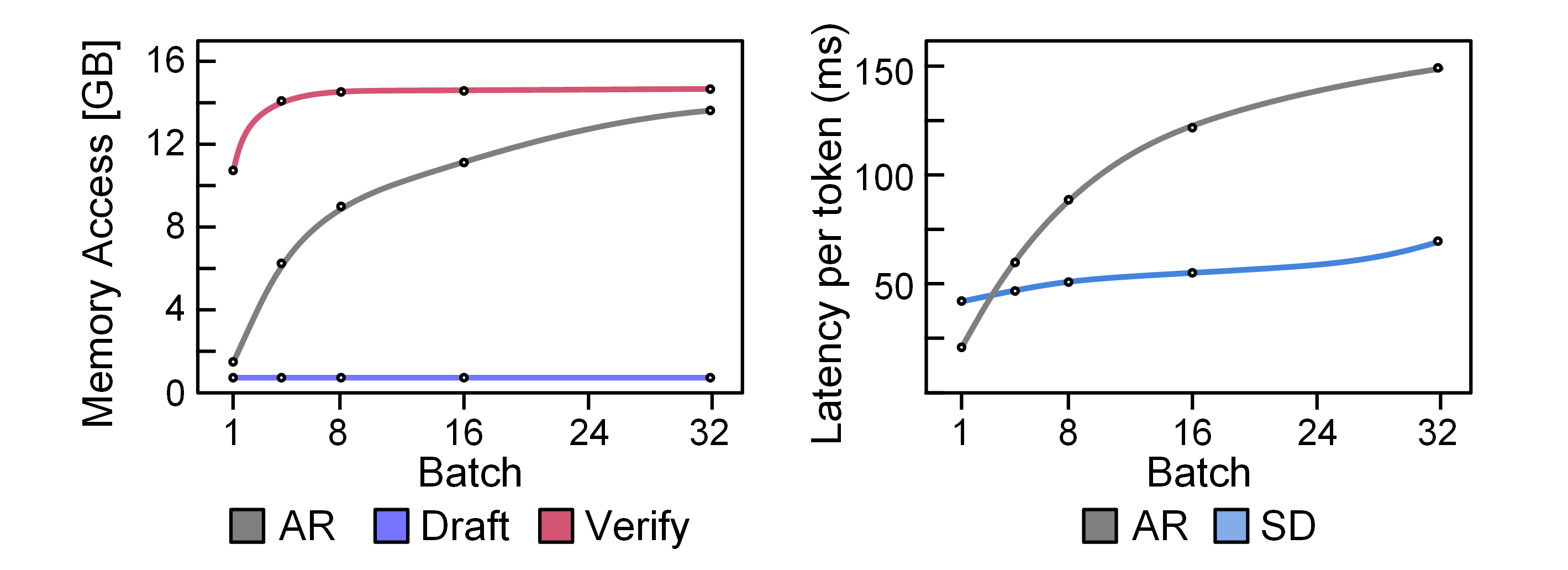}
        \caption{}
        \label{fig_4_b}
    \end{subfigure}
    \caption{(a) Arithmetic intensity of MoE serving (DeepSeek-V2-Lite). (b) Phase-wise memory access and latency per accepted token for AR and EAGLE3-based SD on DGX Spark.}
    \label{fig_4}
    \vspace{-0.5cm}
\end{figure}
}

\newcommand{\figfive}{
\begin{figure}[t]
    \centering
    \begin{subfigure}{\linewidth}
        \centering
        \includegraphics[width=\linewidth]{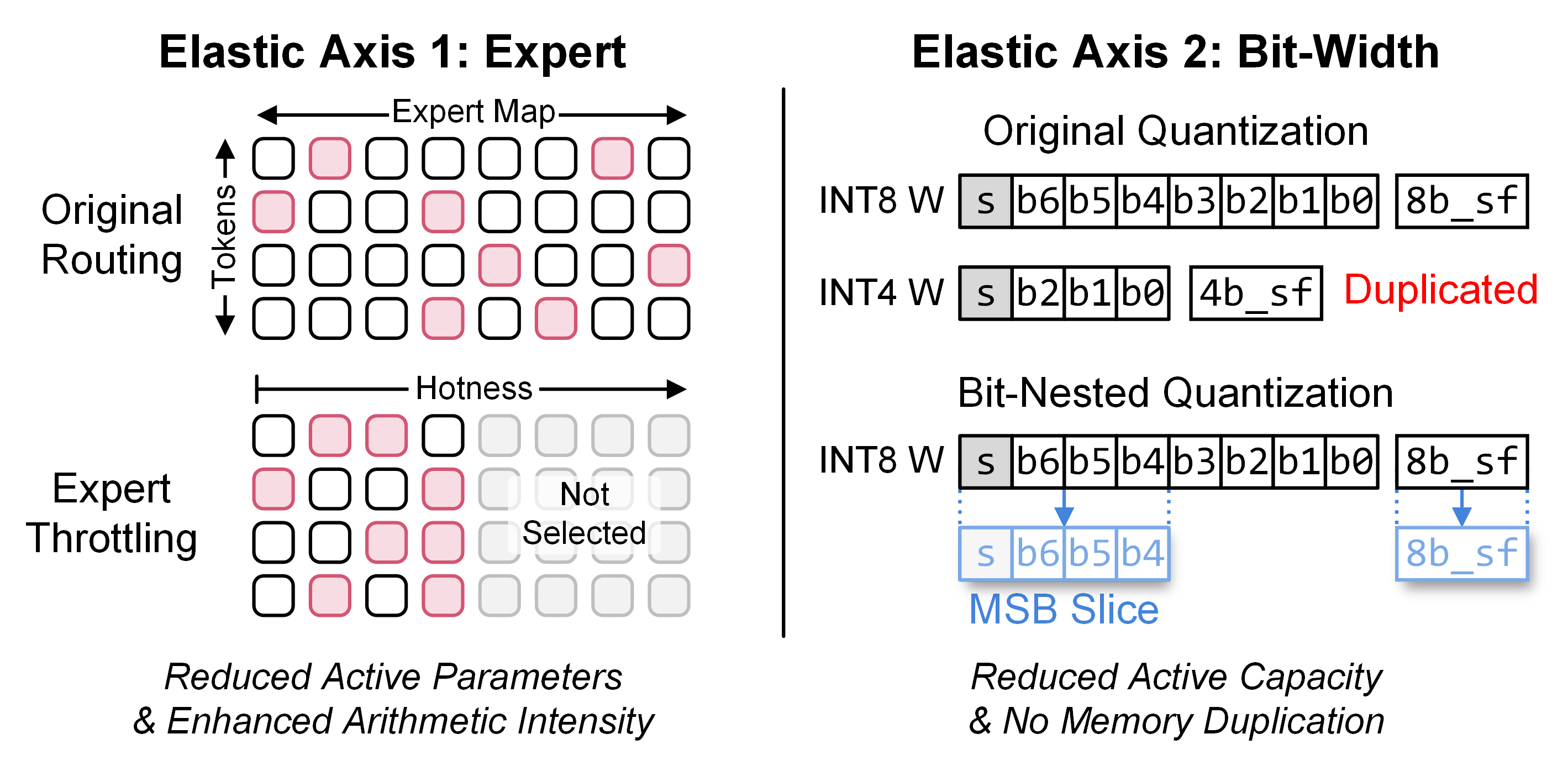}
        \caption{}
        \label{fig_5_a}
    \end{subfigure}
    \begin{subfigure}{\linewidth}
        \centering
        \includegraphics[width=\linewidth]{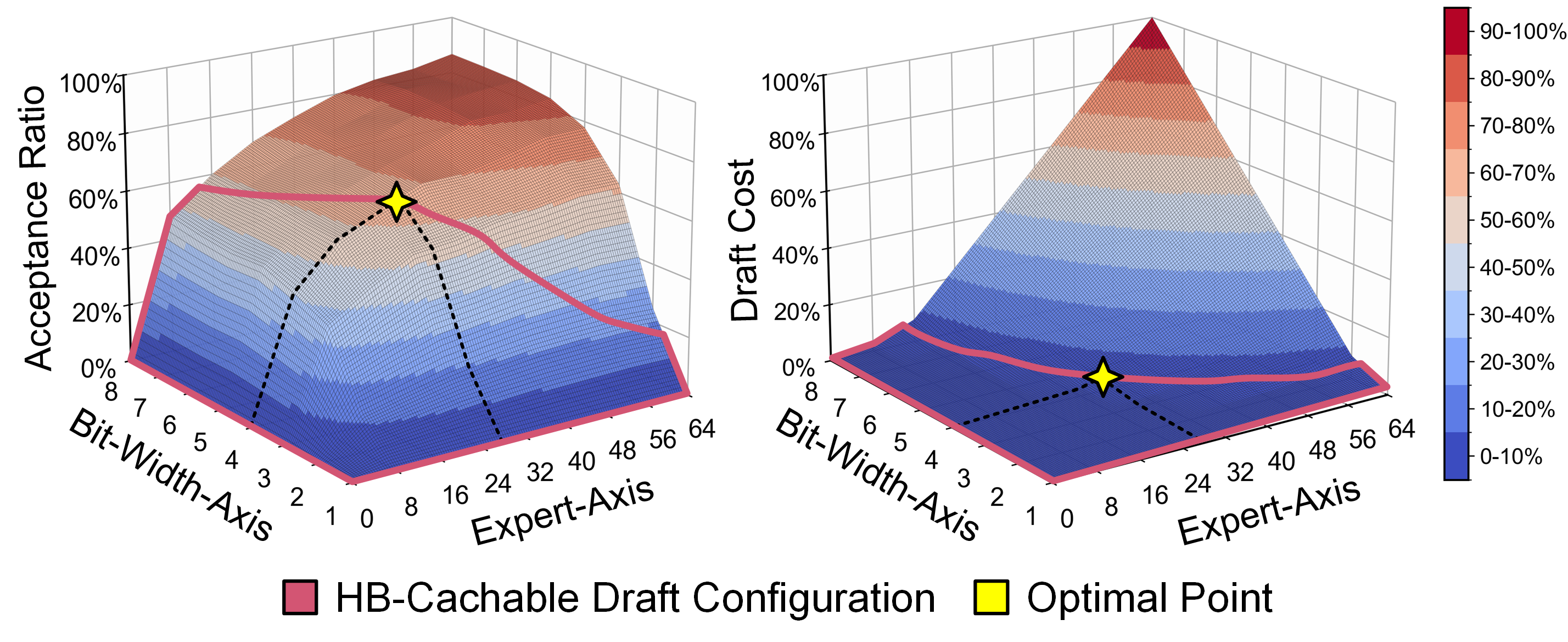}
        \caption{}
        \label{fig_5_b}
    \end{subfigure}
    \caption{(a) Expert and bit-width elasticity axes of MoE. (b) Acceptance ratio and draft cost under joint axis scaling.}
    \label{fig_5}
    \vspace{-0.5cm}
\end{figure}
}

\newcommand{\figsix}{
\begin{figure*}[t]
    \centering
    \includegraphics[width=\linewidth]{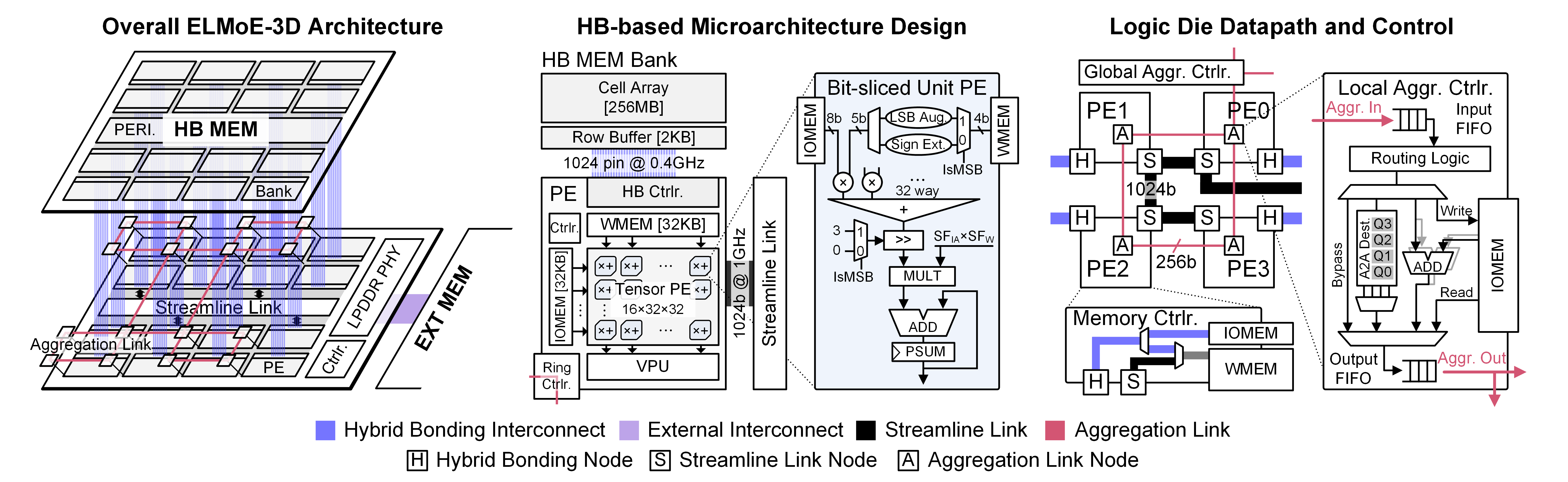}
    \caption{Overall architecture of ELMoE-3D.}
    \label{fig_6}
    \vspace{-0.5cm}
\end{figure*}
}

\newcommand{\figseven}{
\begin{figure}[t]
    \centering
    \includegraphics[width=\linewidth]{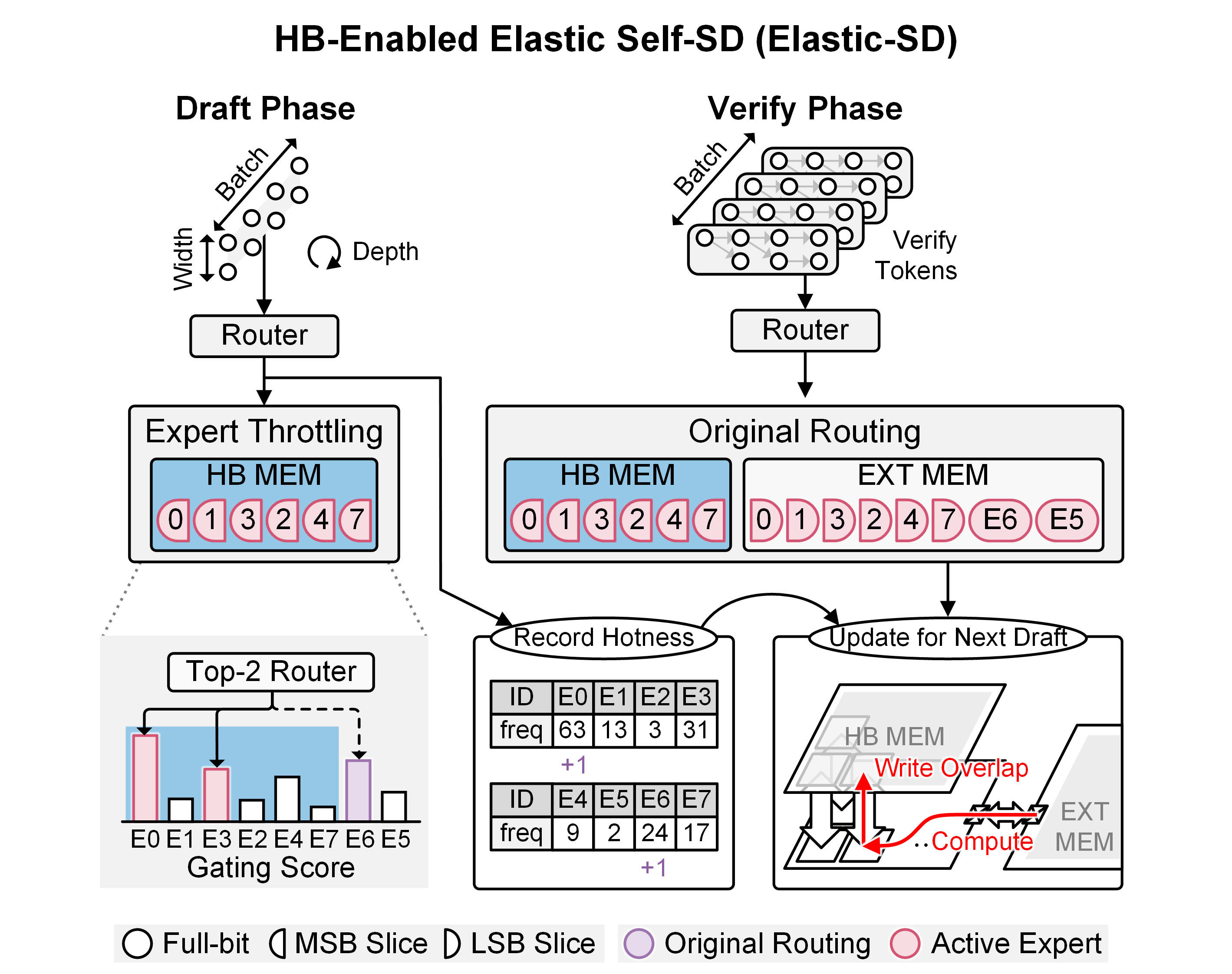}
    \caption{Elastic-SD execution flow.}
    \label{fig_7}
    \vspace{-0.5cm}
\end{figure}
}

\newcommand{\figeight}{
\begin{figure}[t]
    \centering
    \includegraphics[width=\linewidth]{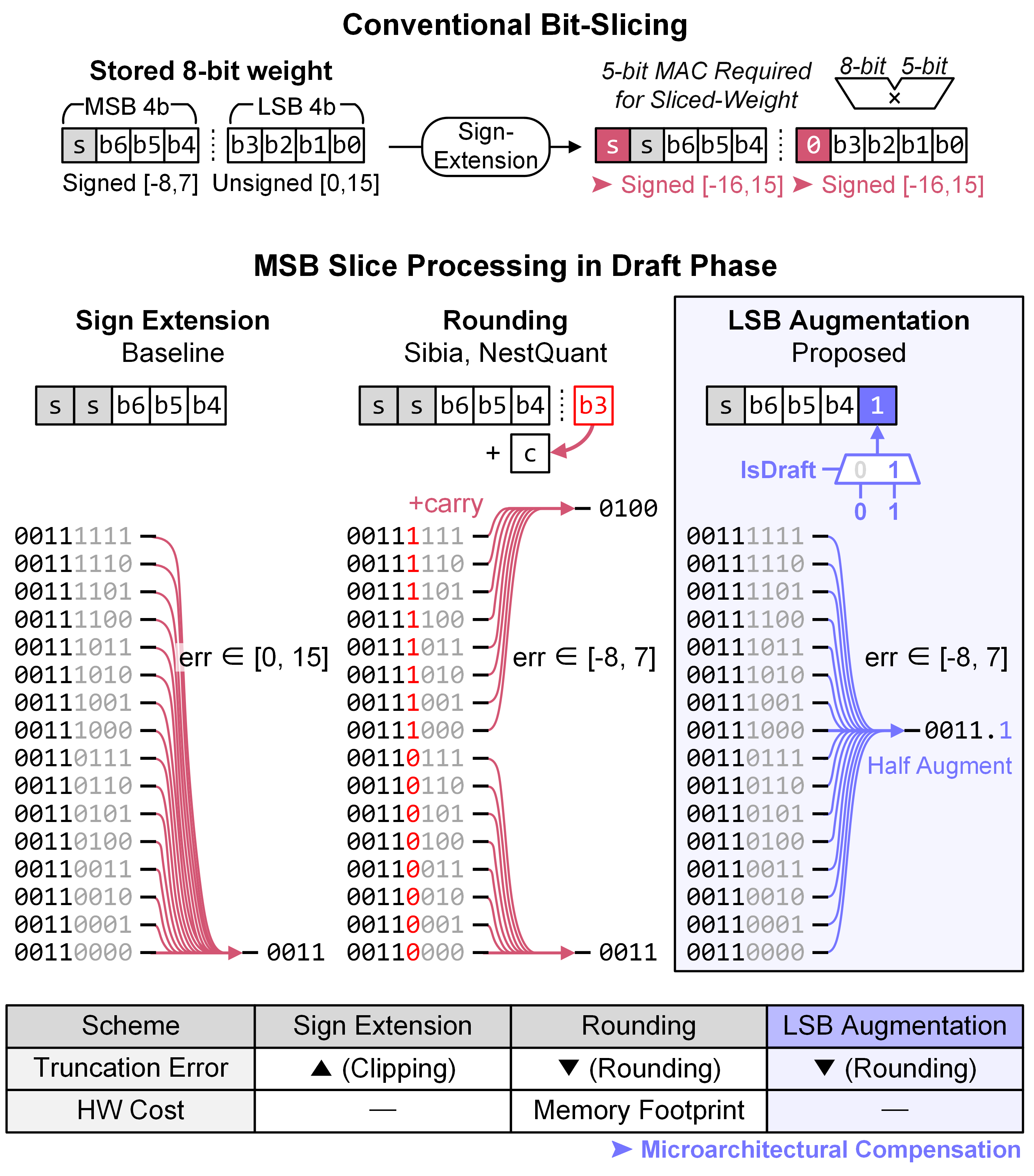}
    \caption{Bit-nested quantization with the proposed LSB augmentation, which provides rounding-like MSB-only draft weights without carry metadata or wider residual slices.}
    \label{fig_8}
    \vspace{-0.5cm}
\end{figure}
}

\newcommand{\fignine}{
\begin{figure}[t]
    \centering
    \begin{subfigure}{\linewidth}
        \centering
        \includegraphics[width=\linewidth]{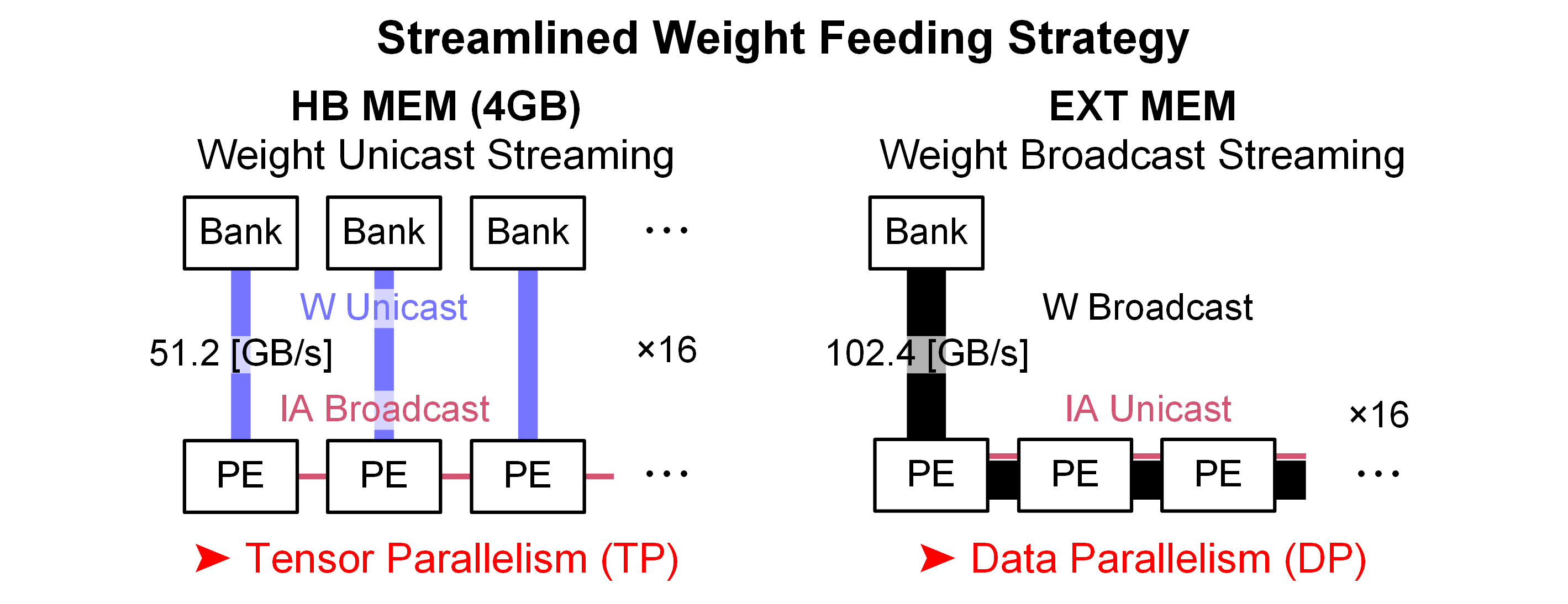}
        \caption{}
        \label{fig_9_a}
    \end{subfigure}
    \begin{subfigure}{\linewidth}
        \centering
        \includegraphics[width=\linewidth]{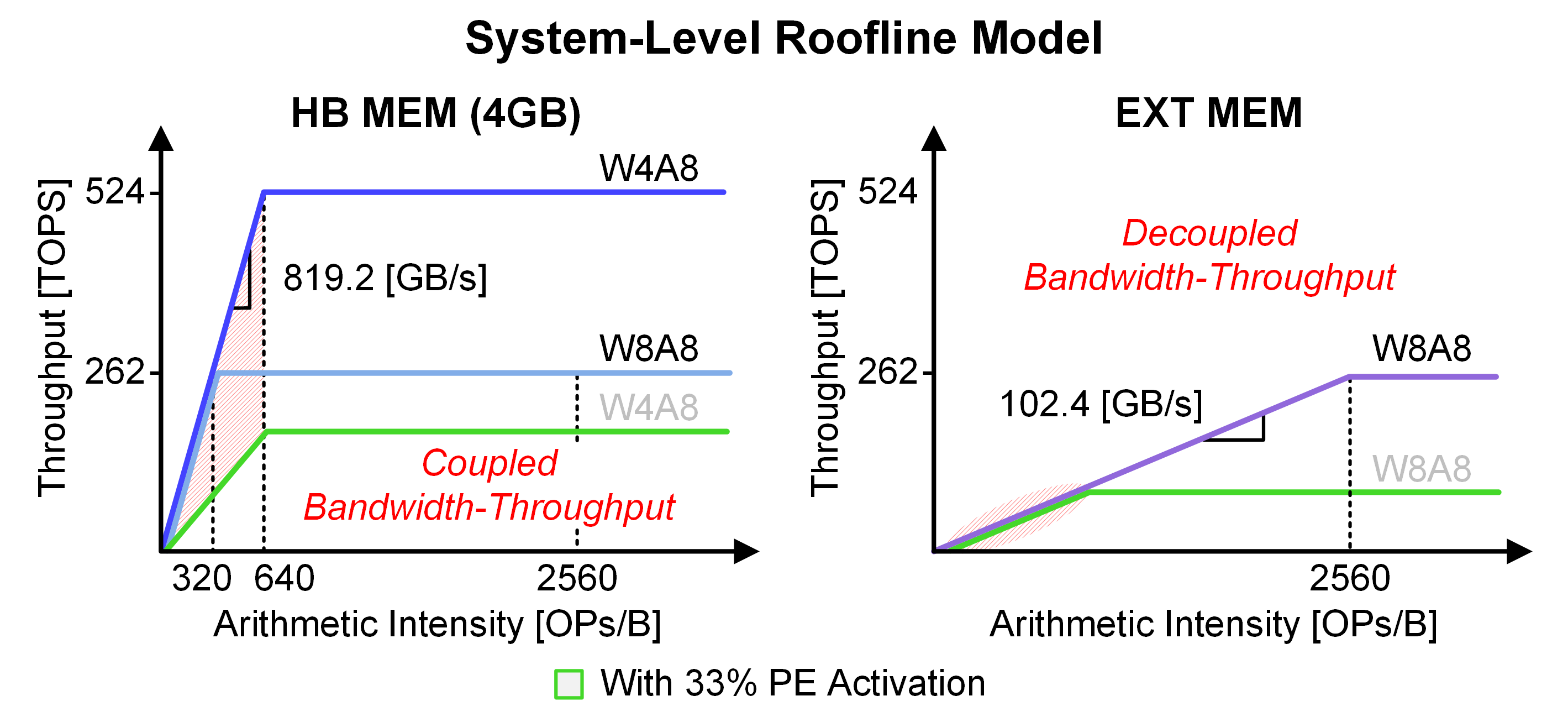}
        \caption{}
        \label{fig_9_b}
    \end{subfigure}
    \caption{(a) Coupled (HB) and decoupled (EXT) bandwidth-throughput characteristics. (b) System-level roofline.}
    \label{fig_9}
    \vspace{-0.5cm}
\end{figure}
}

\newcommand{\figten}{
\begin{figure}[t]
    \centering
    \begin{subfigure}{\linewidth}
        \centering
        \includegraphics[width=\linewidth]{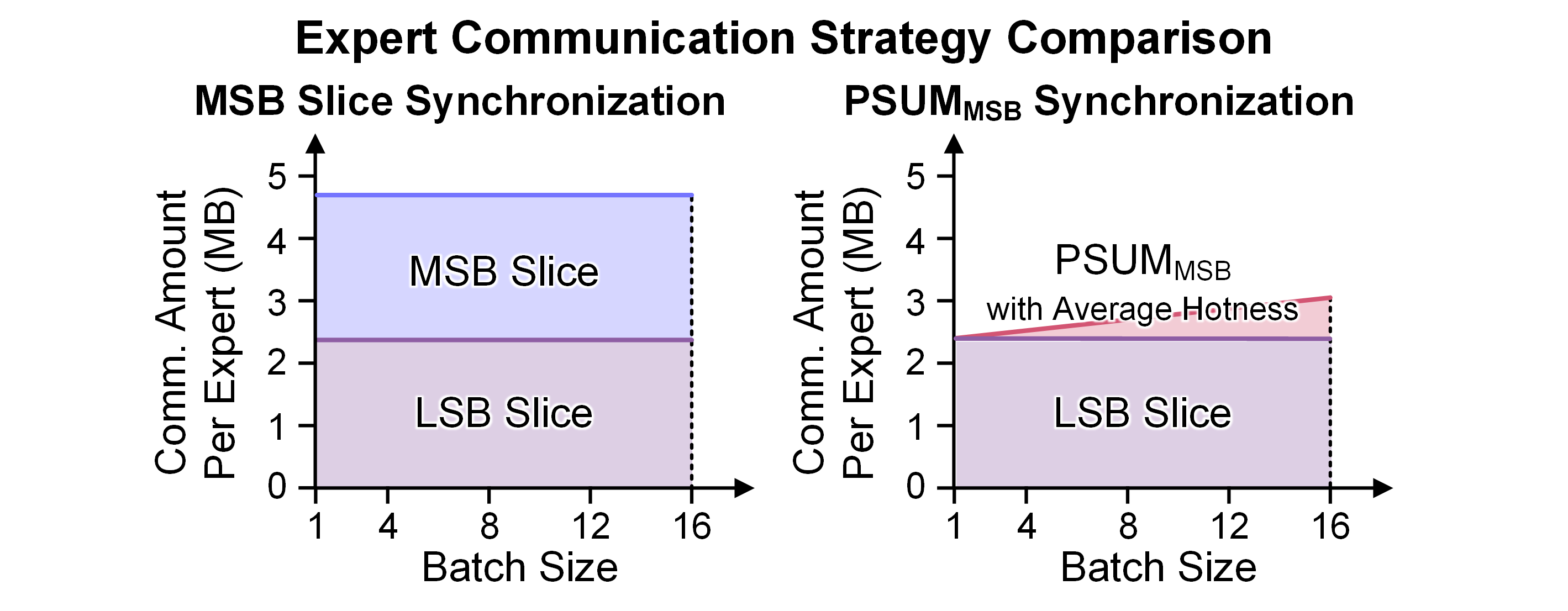}
        \caption{}
        \label{fig_10_a}
    \end{subfigure}
    \begin{subfigure}{\linewidth}
        \centering
        \includegraphics[width=\linewidth]{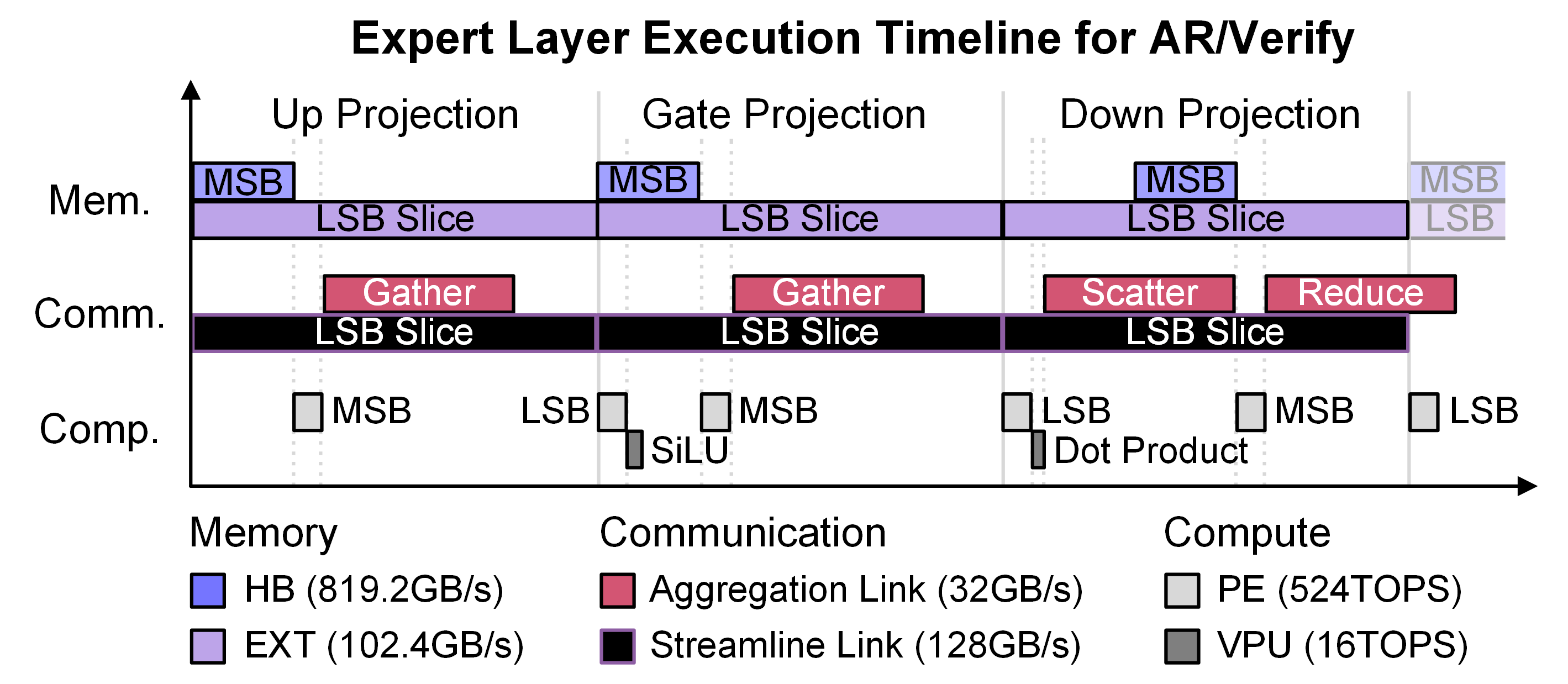}
        \caption{}
        \label{fig_10_b}
    \end{subfigure}
    \caption{(a) On-chip communication volume per expert. (b) Expert-fused execution pipeline.}
    \label{fig_10}
    \vspace{-0.5cm}
\end{figure}
}

\newcommand{\figeleven}{
\begin{figure}[t]
    \centering
    \includegraphics[width=\linewidth]{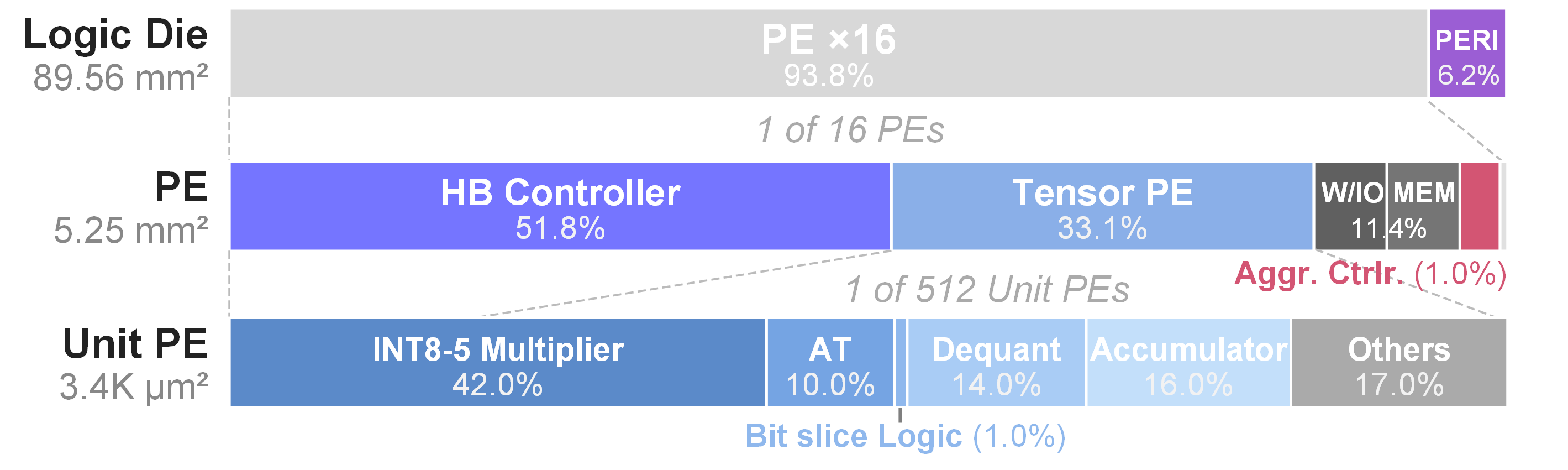}
    \caption{Logic die area breakdown (4GB configuration).}
    \label{fig_11}
    \vspace{-0.5cm}
\end{figure}
}

\newcommand{\figtwelve}{
\begin{figure}[t]
    \centering
    \includegraphics[width=\linewidth]{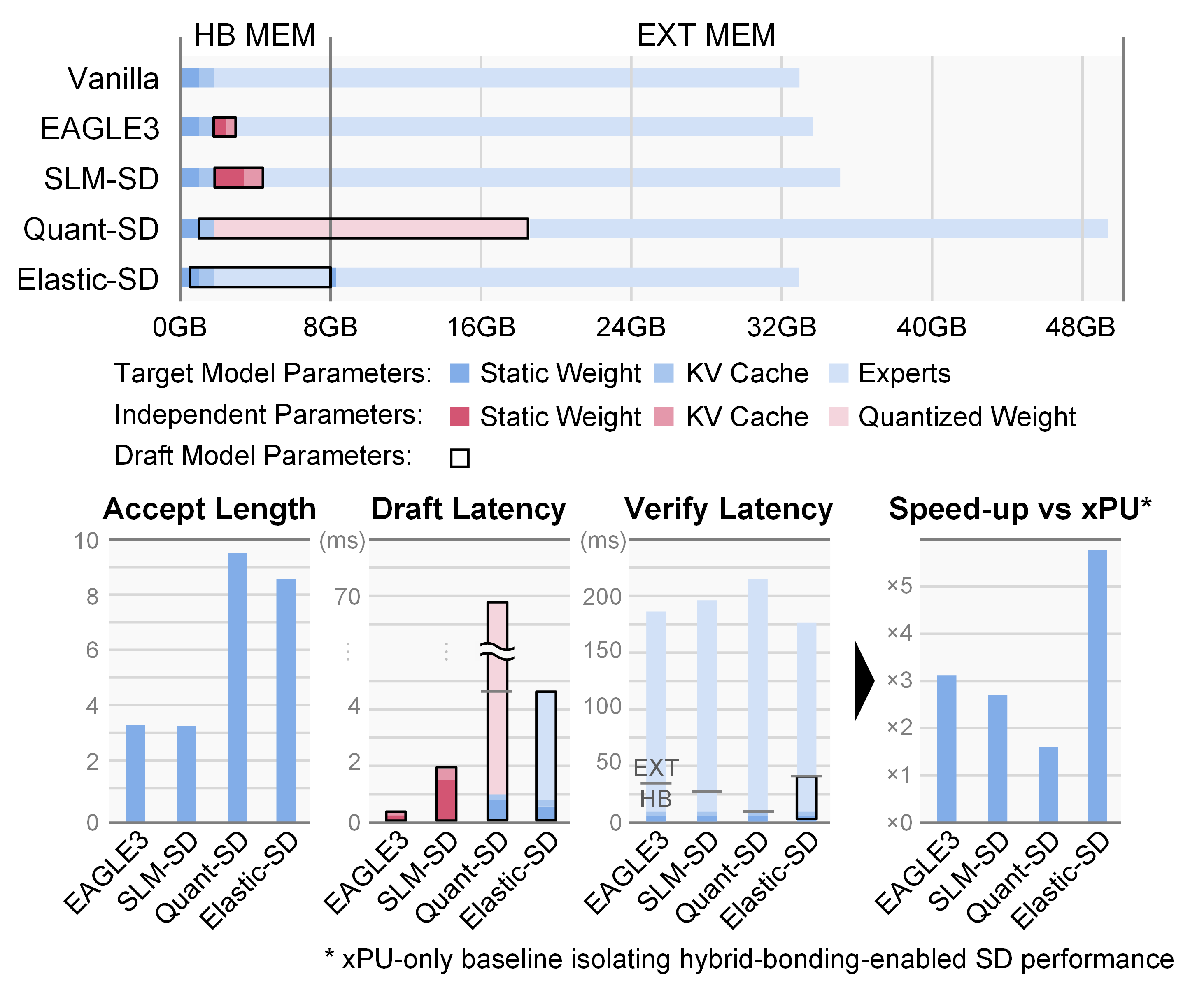}
    \caption{Memory mapping and performance breakdown across SD schemes (Qwen3-30B-A3B, MT-Bench, BS=16, 8GB).}
    \label{fig_12}
    \vspace{-0.5cm}
\end{figure}
}

\newcommand{\figthirteen}{
\begin{figure}[t]
    \centering
    \includegraphics[width=\linewidth]{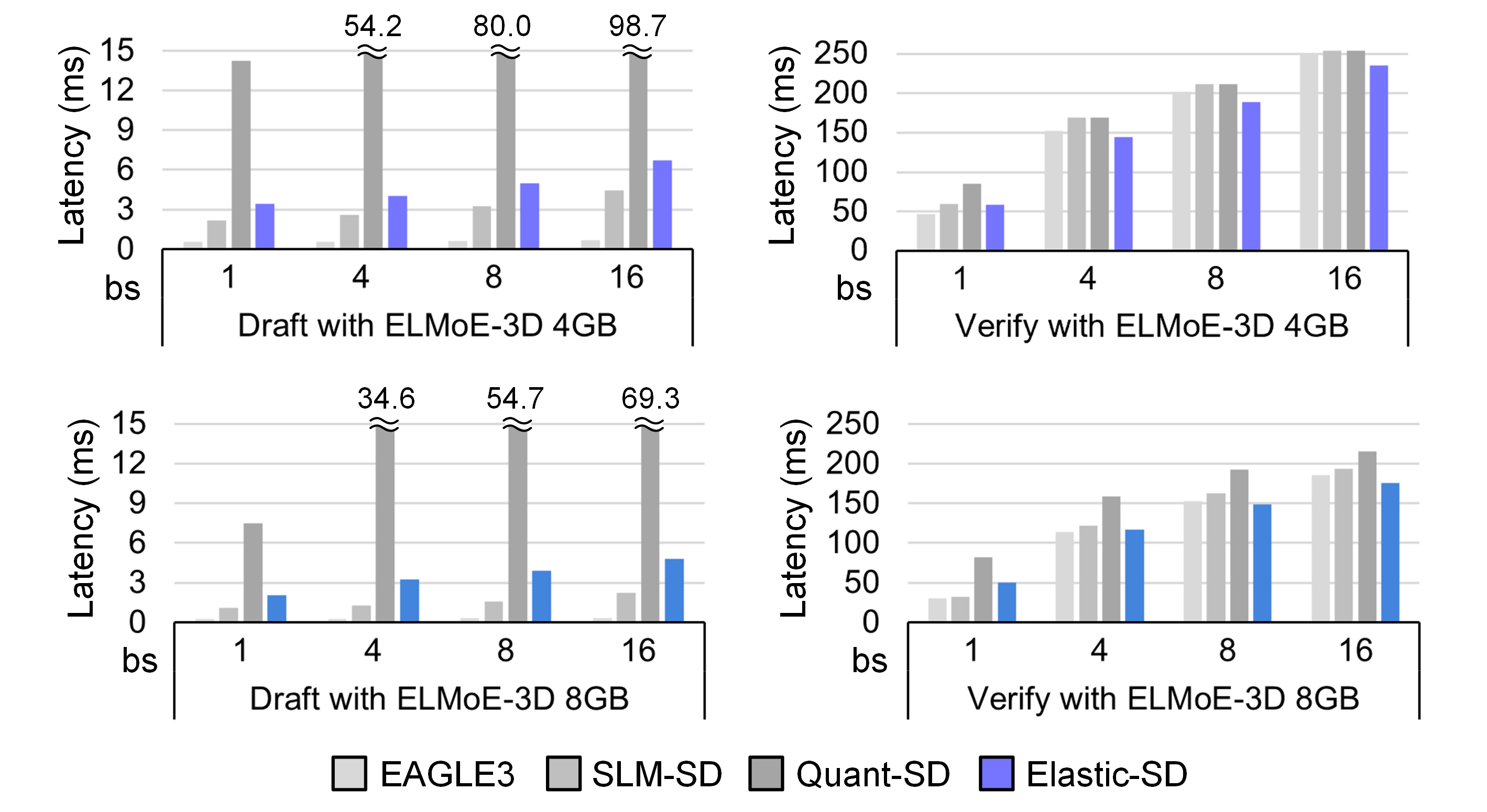}
    \caption{Draft and verify latency across batch sizes and HB capacities (Qwen3-30B-A3B, MT-Bench).}
    \label{fig_13}
    \vspace{-0.5cm}
\end{figure}
}

\newcommand{\figfourteen }{
\begin{figure}[t]
    \centering
    \includegraphics[width=\linewidth]{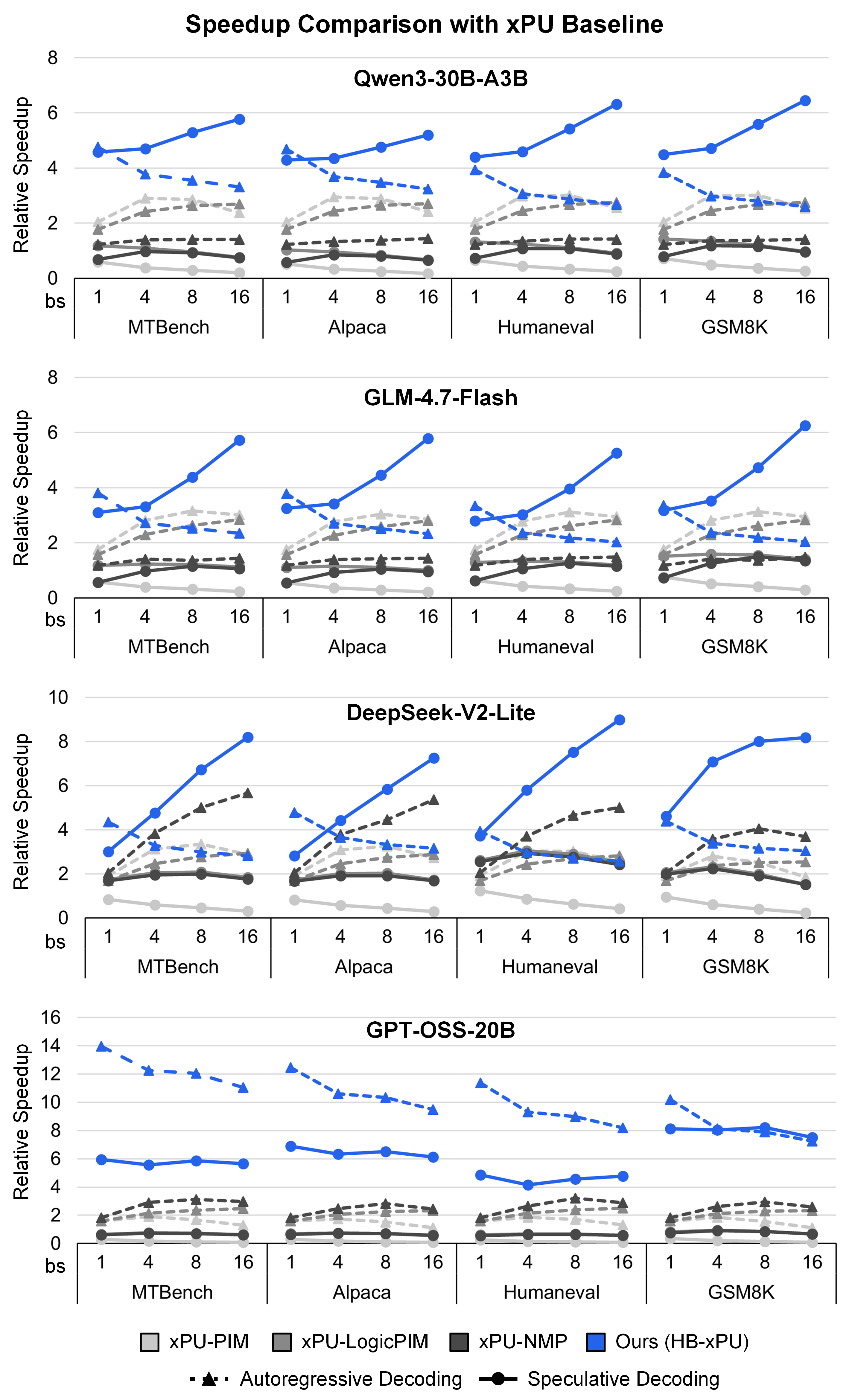}
    \caption{Speedup (AR/SD) over accelerator baselines across models and benchmarks (8GB).}
    \label{fig_14}
    \vspace{-0.5cm}
\end{figure}
}

\newcommand{\figfifteen}{
\begin{figure*}[t]
    \centering
    \includegraphics[width=\linewidth]{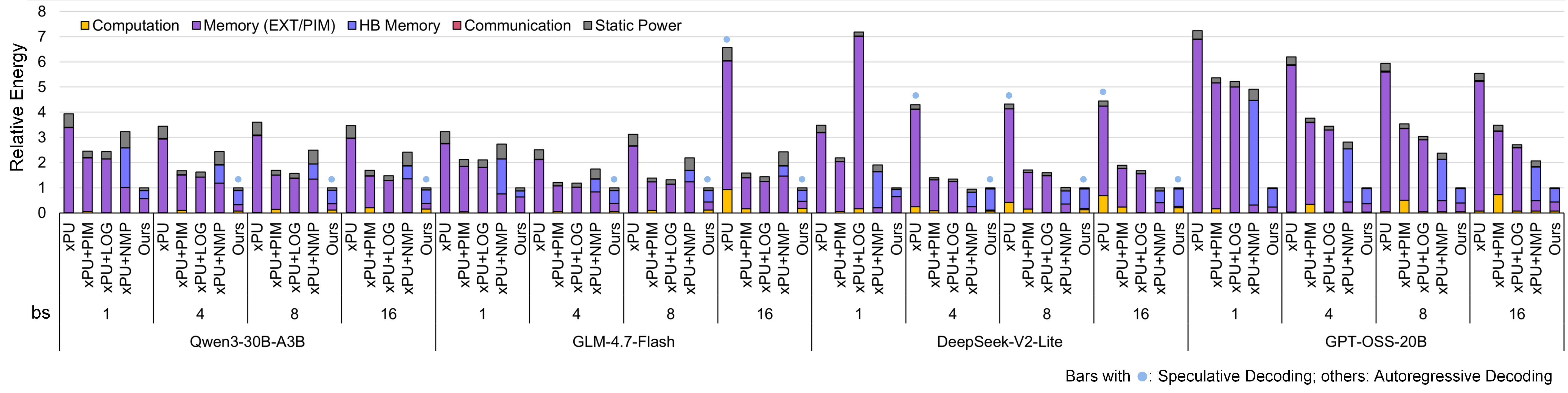}
    \caption{Per-token energy comparison over accelerator baselines normalized to ours across models and batch sizes (MT-Bench, 8GB). Blue markers indicate SD-selected configurations.}
    \label{fig_15}
    \vspace{-0.5cm}
\end{figure*}
}

\newcommand{\figsixteen}{
\begin{figure}[t]
    \centering
    \includegraphics[width=\linewidth]{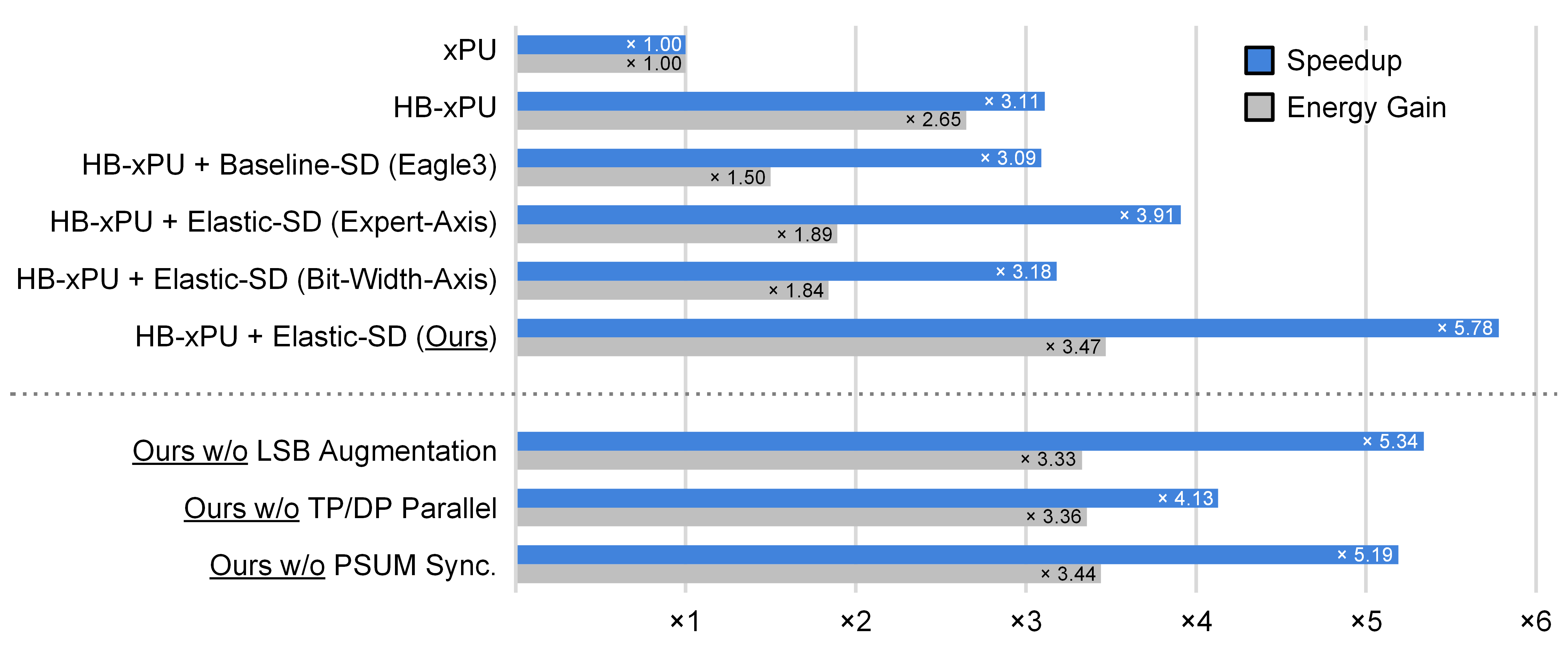}
    \caption{Ablation about speedup and energy gain, showing decomposed contributions and hardware support mechanisms.}
    \label{fig_16}
    \vspace{-0.5cm}
\end{figure}
}

\newcommand{\figseventeen}{
\begin{figure}[t]
    \centering
    \includegraphics[width=\linewidth]{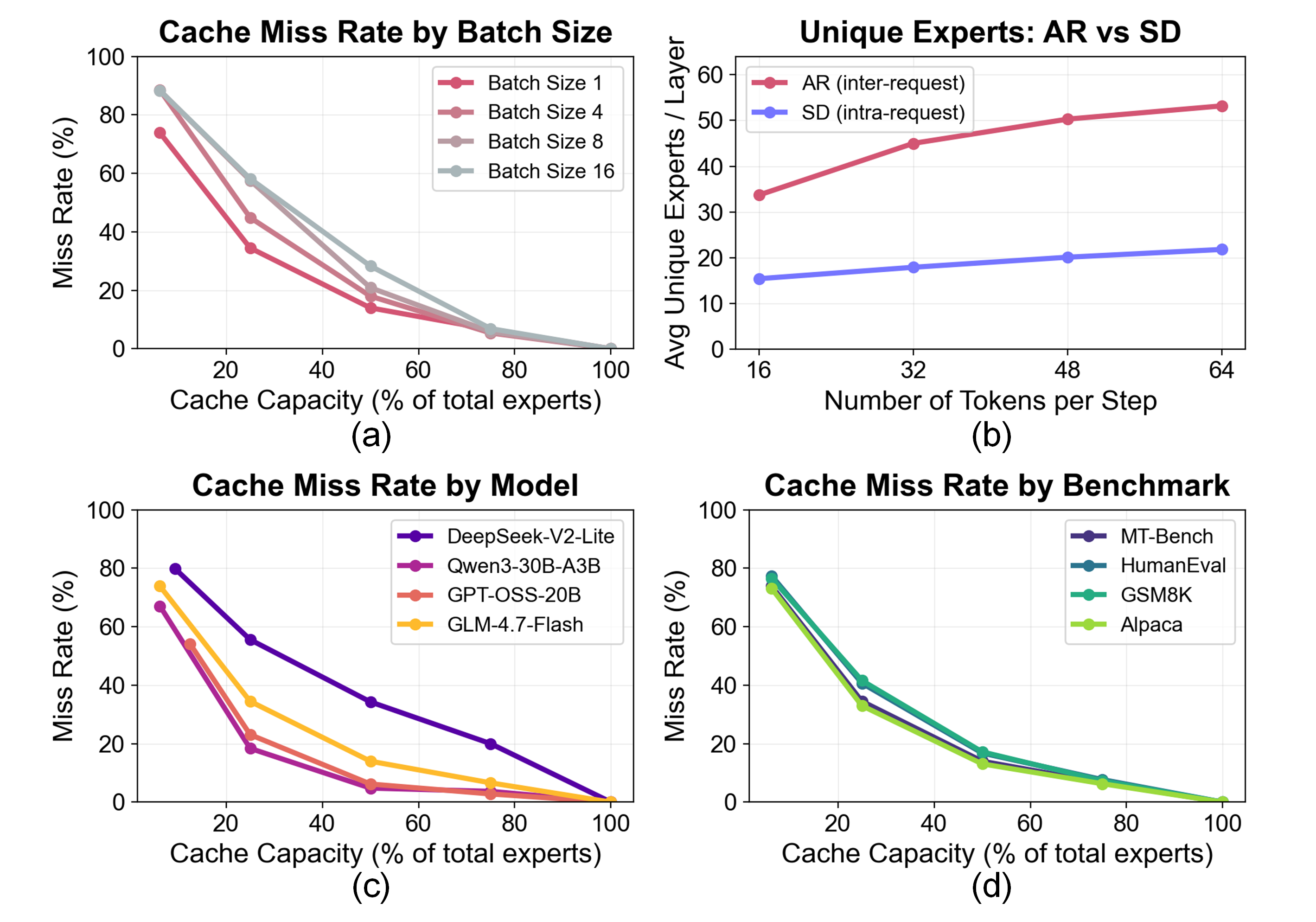}
    \caption{Expert locality analysis across batch size, decoding mode, model, and benchmark (GLM-4.7-Flash, MT-Bench).}
    \label{fig_17}
    \vspace{-0.5cm}
\end{figure}
}

\newcommand{\figeighteen}{
\begin{figure}[t]
    \centering
    \includegraphics[width=\linewidth]{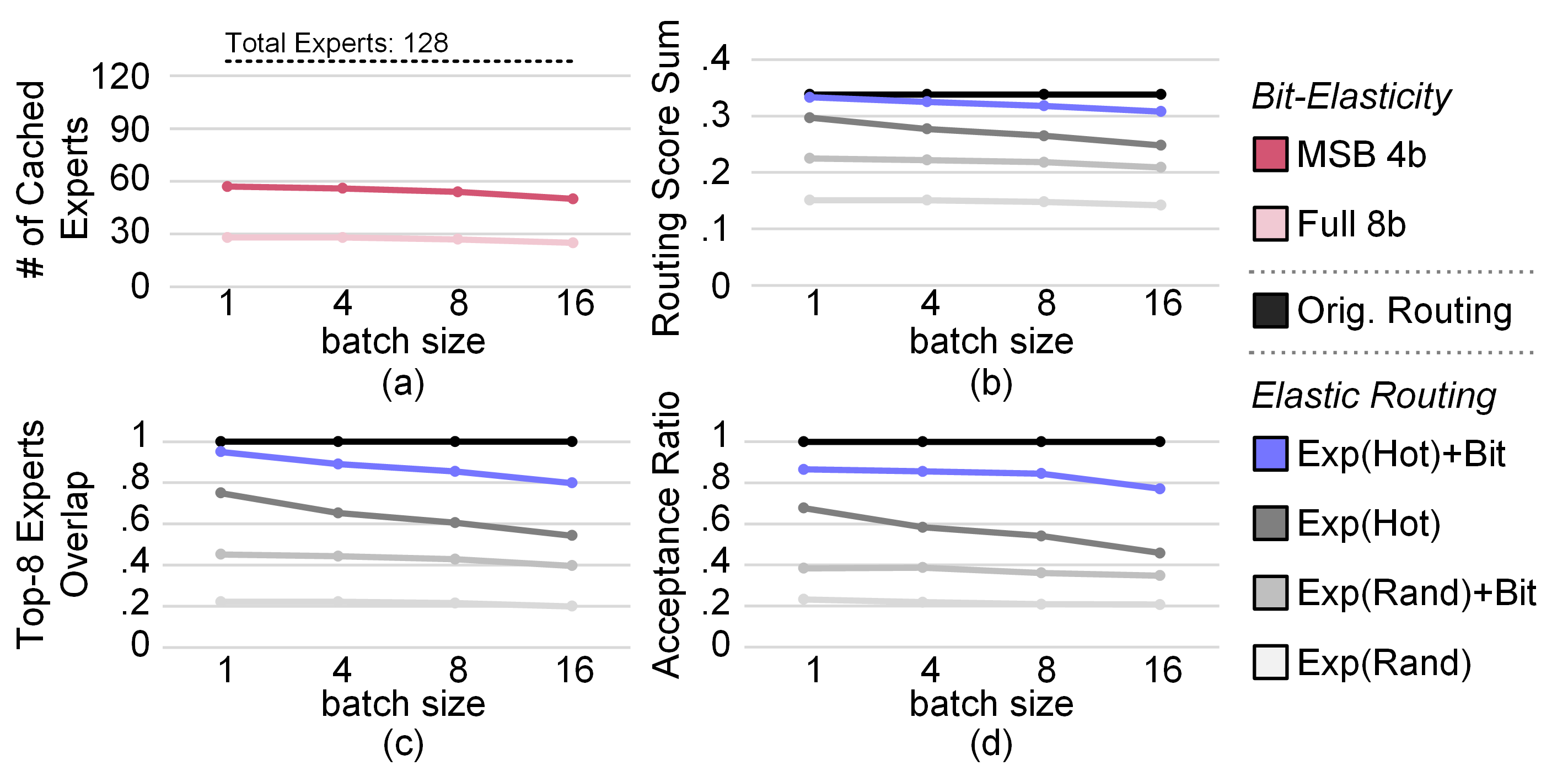}
    \caption{Expert-axis elasticity analysis with hot-expert selection. (Qwen3-30B-A3B, MT-Bench, BS=16, 8GB).}
    \label{fig_18}
    \vspace{-0.5cm}
\end{figure}
}

\newcommand{\fignineteen}{
\begin{figure}[t]
    \centering
    \includegraphics[width=\linewidth]{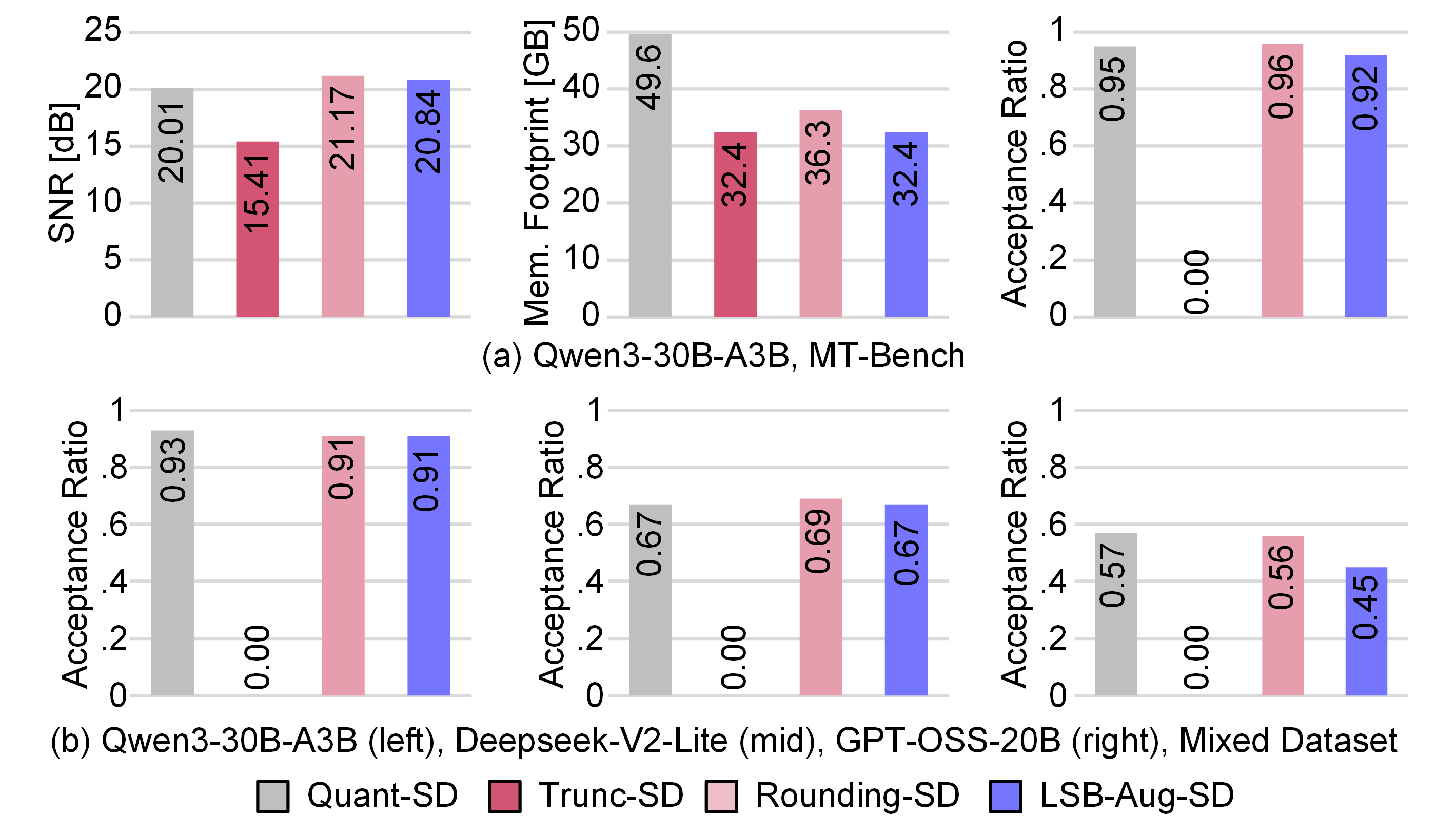}
    \caption{Bitwidth-axis elasticity analysis with LSB augmentation. (Qwen3-30B-A3B, MT-Bench, BS=16, 8GB).}
    \label{fig_19}
    \vspace{-0.5cm}
\end{figure}
}

%% file: Floats/table.tex
\newcommand{\dualcfg}[2]{$\{#1\,|\,#2\}$}

\newcommand{\tabone}{
\begin{table}[t]
\footnotesize
\centering
\setlength{\tabcolsep}{3.2pt}
\caption{ELMoE-3D system specification.}
\label{tab:spec}
\begin{tabular*}{\linewidth}{
@{\extracolsep{\fill}}
@{}l l@{\hspace{0.8em}}
l l@{}
}
\toprule
\multicolumn{2}{c}{\textbf{Logic Die}} &
\multicolumn{2}{c}{\textbf{Memory System}} \\
\cmidrule(r){1-2}
\cmidrule(l{0.02em}){3-4}
\textbf{Parameter} &
\textbf{Value} &
\textbf{Parameter} &
\textbf{Value} \\
\midrule
Technology        & ASAP7 7\,nm~\cite{asap7}
& \textbf{HB MEM}  & \\
Volt./Freq.           & 0.7\,V / 1.0\,GHz
& \quad Banks      & \dualcfg{16}{32} \\
Num PEs           & \dualcfg{16}{32}
& \quad Capacity   & \dualcfg{4}{8}\,GB \\
MACs/PE           & $16{\times}32{\times}32$
& \quad Bandwidth  & \dualcfg{819}{1638}\,GB/s \\
Throughput &
& \quad R/W Energy & 0.43\,pJ/bit~\cite{hb_r_ener} \\
\quad W8A8        & \dualcfg{262}{524}\,TOPS
& \textbf{EXT MEM} & \\
\quad W4A8        & \dualcfg{524}{1048}\,TOPS
& \quad Type       & LPDDR5-6400~\cite{lpddr5_pim} \\
Interconnect &
& \quad Capacity   & 64\,GB \\
\quad Aggreg.     & 32\,GB/s
& \quad Bandwidth  & 102.4\,GB/s \\
\quad Streamline  & 128\,GB/s
& \quad R/W Energy & 3.88\,pJ/bit~\cite{lpddr5_pim} \\
\bottomrule
\end{tabular*}
\vspace{-2mm}
\end{table}
}

\newcommand{\tabtwo}{
\begin{table}[t]
\setlength{\tabcolsep}{4pt}
\footnotesize
\centering
\caption{Target MoE models for on-premises deployment.}
\label{tab_1}
\begin{tabular*}{\linewidth}{@{\extracolsep{\fill}} l cc ccc c}
\toprule
\multirow{2}{*}{\textbf{Model}} 
& \multicolumn{2}{c}{\textbf{Params}} 
& \multicolumn{3}{c}{\textbf{Experts}} 
& \multirow{2}{*}{\textbf{Notes}} \\
\cmidrule(lr){2-3} \cmidrule(lr){4-6}
& \textbf{Tot.} & \textbf{Act.} & \textbf{Tot.} & \textbf{Shr.} & \textbf{Top-k.} & \\
\midrule
Qwen3-30B-A3B    & 30B   & 3B   & 128 & -- & 8 & -- \\
GLM-4.7-Flash    & 30B   & 3B   & 64  & 1  & 4 & -- \\
DeepSeek-V2-Lite & 15.7B & 2.4B & 64  & 2  & 6 & -- \\
GPT-OSS-20B      & 20B   & 3.6B & 32  & -- & 4 & MXFP4 experts \\
\bottomrule
\end{tabular*}
\end{table}
}

\newcommand{\tabthree}{
\begin{table}[t]
\centering
\footnotesize
\setlength{\tabcolsep}{2.0pt}
\caption{Accelerator baseline specifications.}
\label{tab:accel_baselines}
\begin{tabular*}{\linewidth}{@{\extracolsep{\fill}} lccccccc}
\toprule
\multirow{2}{*}{\textbf{Baseline}} &
\textbf{Main} &
\multicolumn{2}{c}{\textbf{HB Mem.}} &
\multicolumn{4}{c}{\textbf{LPDDR5-based EXT Mem.}} \\
\cmidrule(lr){2-2}
\cmidrule(lr){3-4}
\cmidrule(lr){5-8}
&
\textbf{TOPS} &
\textbf{Cap.} &
\textbf{BW} &
\textbf{Cap.} &
\textbf{Ext. BW} &
\textbf{Int. BW} &
\textbf{TOPS} \\
&
&
\textbf{(GB)} &
\textbf{(GB/s)} &
\textbf{(GB)} &
\textbf{(GB/s)} &
\textbf{(GB/s)} &
\\
\midrule
xPU &
262/512 &
-- &
-- &
64 &
102.4 &
-- &
-- \\

\hspace{0.35em}+ PIM~\cite{lpddr5_pim} &
262/512 &
-- &
-- &
64 &
102.4 &
819.2 &
1.64 \\

\hspace{0.35em}+ LogicPIM~\cite{duplex} &
262/512 &
-- &
-- &
64 &
102.4 &
409.6 &
6.55 \\

\multirow{2}{*}{\hspace{0.35em}+ NMP~\cite{h2llm}} &
\multirow{2}{*}{262/512} &
\multirow{2}{*}{--} &
\multirow{2}{*}{--} &
48 &
76.8 &
-- &
-- \\
&
&
&
&
16 &
25.6 &
1638.4 &
13.1 \\

\multirow{2}{*}{HB-xPU} &
262 &
4 &
819.2 &
\multirow{2}{*}{64} &
\multirow{2}{*}{102.4} &
\multirow{2}{*}{--} &
\multirow{2}{*}{--} \\
&
512 &
8 &
1638.4 &
&
&
&
\\
\bottomrule
\end{tabular*}
\end{table}
}

\newcommand{\tabfour}{
\begin{table*}[t]
\centering
\footnotesize
\setlength{\tabcolsep}{6pt}
\caption{Acceptance ratio ($\alpha$) and speedup ($\Phi_{\text{AR}}$, $\Phi_{\text{SD}}$) across draft types on MT-Bench. \textbf{Bold} indicates the best per scenario.}
\label{tab_3}
\begin{threeparttable}
\begin{tabular}{llcccccccccccc}
\toprule
\multirow{2}{*}{Target} & \multirow{2}{*}{Draft}
& \multicolumn{3}{c}{batch-size 1}
& \multicolumn{3}{c}{batch-size 4}
& \multicolumn{3}{c}{batch-size 8}
& \multicolumn{3}{c}{batch-size 16}
\\
\cmidrule(lr){3-5} \cmidrule(lr){6-8}
\cmidrule(lr){9-11} \cmidrule(lr){12-14}
&
& $\alpha$ & $\Phi_{\text{AR}}$ & $\Phi_{\text{SD}}$
& $\alpha$ & $\Phi_{\text{AR}}$ & $\Phi_{\text{SD}}$
& $\alpha$ & $\Phi_{\text{AR}}$ & $\Phi_{\text{SD}}$
& $\alpha$ & $\Phi_{\text{AR}}$ & $\Phi_{\text{SD}}$
\\
\midrule

\multicolumn{14}{c}{\textbf{Hybrid-Bonded Memory = 4GB}} \\
\midrule

\multirow{4}{*}{Qwen3-30B-A3B~\cite{qwen3}}
& EAGLE3~\cite{eagle3}   & 0.33 & 2.83 & 2.66 & 0.33 & 2.24 & 1.89 & 0.33 & 2.06 & 2.07 & 0.33 & 1.83 & \textbf{2.28} \\
& SLM-SD~\cite{slm_sd_alignment}     & 0.33 & 2.21 & 1.70 & 0.33 & 1.23 & 1.53 & 0.33 & 1.17 & 1.76 & 0.33 & 1.16 & 1.96 \\
& Quant-SD~\cite{subspec}   & 0.95 & 1.56 & 1.48 & 0.95 & 1.23 & 1.02 & 0.95 & 1.17 & 1.04 & 0.95 & 1.16 & 1.15 \\
\rowcolor[HTML]{E0F2F5} \cellcolor{white} & Elastic-SD & 0.66 & \textbf{3.08} & 2.64 & 0.57 & 2.48 & \textbf{2.52} & 0.48 & 2.31 & \textbf{2.42} & 0.36 & 2.13 & 2.05 \\
\midrule

\multirow{4}{*}{GLM-4.7-Flash~\cite{glm_4_5}}
& EAGLE3~\cite{eagle3}   & 0.25 & 2.50 & 1.52 & 0.25 & 1.83 & 1.29 & 0.25 & 1.60 & 1.67 & 0.25 & 1.55 & 2.17 \\
& SLM-SD~\cite{slm_sd_alignment}     & 0.14 & 1.79 & 0.60 & 0.14 & 1.27 & 0.74 & 0.14 & 1.18 & 1.00 & 0.14 & 1.13 & 1.27 \\
& Quant-SD~\cite{subspec}   & 0.83 & 1.79 & 0.87 & 0.83 & 1.27 & 0.65 & 0.83 & 1.18 & 0.80 & 0.83 & 1.13 & 1.03 \\
\rowcolor[HTML]{E0F2F5} \cellcolor{white} & Elastic-SD & 0.61 & \textbf{2.61} & 2.02 & 0.55 & 1.93 & \textbf{2.10} & 0.50 & 1.71 & \textbf{2.58} & 0.47 & 1.64 & \textbf{3.13} \\
\midrule

\multirow{4}{*}{DeepSeek-V2-Lite~\cite{deepseek_v2}}
& EAGLE3~\cite{eagle3}   & 0.42 & 2.45 & \textbf{2.63} & 0.42 & 1.87 & 2.80 & 0.42 & 1.71 & 3.65 & 0.42 & 1.62 & 4.21 \\
& SLM-SD~\cite{slm_sd_alignment}     & 0.38 & 1.98 & 1.42 & 0.38 & 1.48 & 1.86 & 0.38 & 1.34 & 2.45 & 0.38 & 1.24 & 2.78 \\
& Quant-SD~\cite{subspec}   & 0.84 & 1.65 & 0.97 & 0.84 & 1.22 & 0.94 & 0.84 & 1.15 & 1.26 & 0.84 & 1.12 & 1.53 \\
\rowcolor[HTML]{E0F2F5} \cellcolor{white} & Elastic-SD & 0.72 & 2.51 & 2.02 & 0.72 & 1.92 & \textbf{3.23} & 0.70 & 1.82 & \textbf{4.34} & 0.70 & 1.72 & \textbf{5.16} \\
\midrule

\multirow{4}{*}{GPT-OSS-20B~\cite{gpt_oss}}
& EAGLE3~\cite{eagle3}   & 0.23 & 4.14 & 2.98 & 0.23 & 2.90 & 1.98 & 0.23 & 2.78 & 2.13 & 0.23 & 1.22 & 2.13 \\
& SLM-SD~\cite{slm_sd_alignment}     & 0.24 & 1.81 & 1.29 & 0.24 & 1.33 & 1.54 & 0.24 & 1.25 & 1.71 & 0.24 & 1.22 & 1.74 \\
& Quant-SD~\cite{subspec}   & 0.60 & 5.54 & 1.65 & 0.60 & 4.85 & 0.74 & 0.60 & 4.36 & 0.69 & 0.60 & 3.94 & 0.74 \\
\rowcolor[HTML]{E0F2F5} \cellcolor{white} & Elastic-SD & 0.44 & \textbf{5.54} & 3.08 & 0.39 & \textbf{4.85} & 2.86 & 0.36 & \textbf{4.36} & 2.78 & 0.33 & \textbf{3.94} & 2.56 \\

\midrule

\multicolumn{14}{c}{\textbf{Hybrid-Bonded Memory = 8GB}} \\
\midrule

\multirow{4}{*}{Qwen3-30B-A3B~\cite{qwen3}}
& EAGLE3~\cite{eagle3}   & 0.33 & 4.44 & 4.25 & 0.33 & 3.57 & 2.55 & 0.33 & 3.35 & 2.76 & 0.33 & 3.11 & 3.09 \\
& SLM-SD~\cite{slm_sd_alignment}     & 0.33 & 4.11 & 3.21 & 0.33 & 3.22 & 2.21 & 0.33 & 2.94 & 2.40 & 0.33 & 2.77 & 2.71 \\
& Quant-SD~\cite{subspec}   & 0.95 & 1.63 & 2.15 & 0.95 & 1.25 & 1.44 & 0.95 & 1.19 & 1.42 & 0.95 & 1.17 & 1.56 \\
\rowcolor[HTML]{E0F2F5} \cellcolor{white} & Elastic-SD & 0.91 & \textbf{4.77} & 4.58 & 0.91 & 3.78 & \textbf{4.71} & 0.90 & 3.56 & \textbf{5.29} & 0.86 & 3.31 & \textbf{5.78} \\
\midrule

\multirow{4}{*}{GLM-4.7-Flash~\cite{glm_4_5}}
& EAGLE3~\cite{eagle3}   & 0.25 & 3.70 & 2.27 & 0.25 & 2.63 & 1.63 & 0.25 & 2.43 & 2.13 & 0.25 & 2.26 & 2.76 \\
& SLM-SD~\cite{slm_sd_alignment}     & 0.14 & 3.34 & 1.24 & 0.14 & 2.36 & 0.98 & 0.14 & 2.09 & 1.27 & 0.14 & 1.93 & 1.63 \\
& Quant-SD~\cite{subspec}   & 0.83 & 1.90 & 1.30 & 0.83 & 1.30 & 0.90 & 0.83 & 1.19 & 1.08 & 0.83 & 1.14 & 1.38 \\
\rowcolor[HTML]{E0F2F5} \cellcolor{white} & Elastic-SD & 0.78 & \textbf{3.82} & 3.10 & 0.76 & 2.73 & \textbf{3.31} & 0.75 & 2.52 & \textbf{4.39} & 0.74 & 2.34 & \textbf{5.74} \\
\midrule

\multirow{4}{*}{DeepSeek-V2-Lite~\cite{deepseek_v2}}
& EAGLE3~\cite{eagle3}   & 0.42 & 4.22 & \textbf{4.95} & 0.42 & 3.08 & 4.38 & 0.42 & 2.90 & 5.71 & 0.42 & 2.74 & 6.59 \\
& SLM-SD~\cite{slm_sd_alignment}     & 0.38 & 3.46 & 2.87 & 0.38 & 2.56 & 3.01 & 0.38 & 2.34 & 3.86 & 0.38 & 2.21 & 4.40 \\
& Quant-SD~\cite{subspec}   & 0.84 & 1.73 & 1.73 & 0.84 & 1.24 & 2.03 & 0.84 & 1.16 & 2.64 & 0.84 & 1.13 & 3.18 \\
\rowcolor[HTML]{E0F2F5} \cellcolor{white} & Elastic-SD & 0.84 & 4.35 & 3.01 & 0.83 & 3.28 & \textbf{4.76} & 0.83 & 2.99 & \textbf{6.72} & 0.82 & 2.83 & \textbf{8.20} \\
\midrule

\multirow{4}{*}{GPT-OSS-20B~\cite{gpt_oss}}
& EAGLE3~\cite{eagle3}   & 0.23 & 10.70 & 7.13 & 0.23 & 9.27 & 4.36 & 0.23 & 9.00 & 4.27 & 0.23 & 8.26 & 4.08 \\
& SLM-SD~\cite{slm_sd_alignment}     & 0.24 & 10.14 & 4.71 & 0.24 & 8.11 & 3.38 & 0.24 & 7.82 & 3.35 & 0.24 & 6.68 & 3.10 \\
& Quant-SD~\cite{subspec}   & 0.60 & 13.95 & 4.42 & 0.60 & 12.24 & 1.60 & 0.60 & 12.03 & 1.47 & 0.60 & 11.05 & 1.48 \\
\rowcolor[HTML]{E0F2F5} \cellcolor{white} & Elastic-SD & 0.50 & \textbf{13.95} & 5.95 & 0.50 & \textbf{12.24} & 5.57 & 0.50 & \textbf{12.03} & 5.85 & 0.47 & \textbf{11.05} & 5.65 \\

\bottomrule
\end{tabular}
\end{threeparttable}
\end{table*}
}

\newcommand{\tabfive}{
\begin{table*}[t]
\centering
\footnotesize
\setlength{\tabcolsep}{3.0pt}
\caption{Acceptance ratio across models, datasets, token types, and positions (Qwen3-30B-A3B, Mixed-set, BS=16, 8GB).}
\label{tab:alignment_sweep}
\begin{tabular*}{\textwidth}{@{\extracolsep{\fill}} lcccccccccccccccc}
\toprule
\multirow{2}{*}{\textbf{Method}} &
\multicolumn{4}{c}{\textbf{(a) Model Sweep}} &
\multicolumn{5}{c}{\textbf{(b) Dataset Sweep}} &
\multicolumn{4}{c}{\textbf{(c) Token Type Sweep}} &
\multicolumn{3}{c}{\textbf{(d) Position Sweep}} \\
\cmidrule(lr){2-5}
\cmidrule(lr){6-10}
\cmidrule(lr){11-14}
\cmidrule(lr){15-17}
& \textbf{QW3} & \textbf{GLM} & \textbf{GPT} & \textbf{DS2}
& \textbf{MTB} & \textbf{ALP} & \textbf{GSM} & \textbf{HE} & \textbf{Mix}
& \textbf{Struct.} & \textbf{Freq.} & \textbf{Rare} & \textbf{Symb.}
& \textbf{Early} & \textbf{Mid.} & \textbf{Late} \\
\midrule
EAGLE3
& 0.28 & 0.27 & 0.24 & 0.43
& 0.33 & 0.27 & 0.42 & 0.39 & 0.28
& 0.31 & 0.27 & 0.25 & 0.39 
& 0.34 & 0.25 & 0.26 \\
SLM-SD
& 0.33 & 0.15 & 0.15 & 0.37
& 0.33 & 0.28 & 0.46 & 0.35 & 0.33
& 0.33 & 0.31 & 0.32 & 0.58
& 0.32 & 0.35 & 0.33 \\
Quant-SD
& 0.95 & 0.80 & 0.57 & 0.84
& 0.95 & 0.90 & 0.99 & 0.99 & 0.95
& 0.95 & 0.95 & 0.94 & 0.99
& 0.98 & 0.94 & 0.92 \\
Elastic-SD
& 0.87 & 0.70 & 0.47 & 0.83
& 0.86 & 0.76 & 0.96 & 0.94 & 0.87
& 0.83 & 0.93 & 0.83 & 0.93
& 0.90 & 0.87 & 0.83 \\
\bottomrule
\end{tabular*}
\vspace{0.5mm}
\begin{flushleft}
\scriptsize
Abbrev.: QW3=Qwen3-30B-A3B, GLM=GLM-4.7-Flash, GPT=GPT-OSS-20B, DS2=DeepSeek-V2-Lite, MTB=MT-Bench, ALP=Alpaca, GSM=GSM8K, HE=HumanEval, \\Mix=mixed dataset; Struct.=structural, Freq.=frequent content, Rare=rare content, Symb.=symbolic/code tokens; Early=0-0.3k / Mid=0.3k-0.7k / Late=0.7k-1k position
\end{flushleft}
\vspace{-2mm}
\end{table*}
}

\newcommand{\tabsix}{
\begin{table}[t]
\centering
\footnotesize
\setlength{\tabcolsep}{1.8pt}
\caption{Measured expert-pool refresh overhead under MT-Bench continuous batching (Qwen3-30B-A3B, 8GB HB).}
\label{tab:scenario_breakdown}
\begin{tabular*}{\linewidth}{@{\extracolsep{\fill}}ccccccc}
\toprule
\multirow{2}{*}{\shortstack{\textbf{Batch}\\\textbf{Size}}} &
\multirow{2}{*}{\shortstack{\textbf{Request}\\\textbf{Arrival}}} &
\multirow{2}{*}{\shortstack{\textbf{Iter.}\\\textbf{Share}}} &
\multicolumn{2}{c}{\textbf{\# Experts (Total: 128)}} &
\multirow{2}{*}{\shortstack{\textbf{Acc.}\\\textbf{Ratio}\\(Per Req.)}} &
\multirow{2}{*}{\shortstack{\textbf{Latency}\\\textbf{Overhead}\\(Per Iter.)}} \\
\cmidrule(lr){4-5}
& & & \textbf{Verify} & \textbf{Update (Ovl.)} & & \\
\midrule
\multirow{4}{*}{4}
& Decode-only & 96.84\% & 85.0  & 8.9 (87\%)  & 0.94 & -- \\
& Multi-turn  & 1.58\%  & 77.5  & 14.5 (68\%) & 0.53 & 3.5\% \\
& New Query   & 1.58\%  & 79.4  & 16.6 (68\%) & 0.99 & 4.0\% \\
\cmidrule(lr){2-7}
& \textbf{Overall} & 100\% & -- & -- & \textbf{0.93} & \textbf{0.12\%} \\
\midrule
\multirow{4}{*}{16}
& Decode-only & 97.27\% & 111.5 & 6.4 (99\%)  & 0.87 & -- \\
& Multi-turn  & 1.37\%  & 101.0 & 21.2 (88\%) & 0.46 & 1.5\% \\
& New Query   & 1.37\%  & 104.9 & 20.2 (91\%) & 0.85 & 1.1\% \\
\cmidrule(lr){2-7}
& \textbf{Overall} & 100\% & -- & -- & \textbf{0.87} & \textbf{0.04\%} \\
\bottomrule
\end{tabular*}
\vspace{-2mm}
\end{table}
}

\newcommand{\tabseven}{
\begin{table}[t]
\centering
\footnotesize
\setlength{\tabcolsep}{1.8pt}
\caption{Expert-pool refresh overhead across decode-length budgets (Qwen3-30B-A3B, mixed datasets, BS=16, 8GB HB).}
\label{tab:churn_stress}
\begin{tabular*}{\linewidth}{@{\extracolsep{\fill}}ccccccc}
\toprule
\multirow{2}{*}{\shortstack{\textbf{Gen.}\\\textbf{Len.}}} &
\multirow{2}{*}{\shortstack{\textbf{Request}\\\textbf{Arrival}}} &
\multirow{2}{*}{\shortstack{\textbf{Iter.}\\\textbf{Share}}} &
\multicolumn{2}{c}{\textbf{\# Experts (Total: 128)}} &
\multirow{2}{*}{\shortstack{\textbf{Acc.}\\\textbf{Ratio}\\(Per Req.)}} &
\multirow{2}{*}{\shortstack{\textbf{Latency}\\\textbf{Overhead}\\(Per Iter.)}} \\
\cmidrule(lr){4-5}
& & & \textbf{Verify} & \textbf{Update (Ovl.)} & & \\
\midrule
\multirow{3}{*}{128}
& Decode-only & 93.10\% & 114.1 & 10.8 (99\%) & 0.81 & -- \\
& New Query   & 6.90\%  & 110.4 & 20.9 (88\%) & 0.77 & 1.5\% \\
\cmidrule(lr){2-7}
& \textbf{Overall} & 100\% & -- & -- & \textbf{0.81} & \textbf{0.09\%} \\
\midrule
\multirow{3}{*}{64}
& Decode-only & 86.67\% & 113.5 & 15.9 (99\%) & 0.82 & -- \\
& New Query   & 13.33\% & 110.8 & 21.2 (88\%) & 0.77 & 1.5\% \\
\cmidrule(lr){2-7}
& \textbf{Overall} & 100\% & -- & -- & \textbf{0.81} & \textbf{0.17\%} \\
\midrule
\multirow{3}{*}{32}
& Decode-only & 76.38\% & 113.4 & 24.5 (98\%) & 0.81 & -- \\
& New Query   & 23.62\% & 110.4 & 18.6 (87\%) & 0.81 & 1.4\% \\
\cmidrule(lr){2-7}
& \textbf{Overall} & 100\% & -- & -- & \textbf{0.81} & \textbf{0.23\%} \\
\bottomrule
\end{tabular*}
\vspace{-2mm}
\end{table}
}

\newcommand{\tabeight}{
\begin{table}[t]
\centering
\footnotesize
\setlength{\tabcolsep}{2.3pt}
\caption{Long-context KV pressure on EAGLE-3 (EG) and Elastic-SD (EL) (Qwen3-30B-A3B, LongBench QMSum, BS=8, 8GB HB). Sparse applies a fixed 4k-token KV attention budget only during drafting; verification uses full attention.}
\label{tab:qmsum}
\begin{tabular*}{\linewidth}{@{\extracolsep{\fill}}llccccccc}
\toprule
\multirow{2}{*}{\shortstack{\textbf{Seq.}\\\textbf{Len.}}} &
\multirow{2}{*}{\shortstack{\textbf{Attn.}\\\textbf{Type}}} &
\multicolumn{2}{c}{\textbf{\# Cached Exps}} &
\multicolumn{2}{c}{\textbf{Acc. Ratio}} &
\multicolumn{2}{c}{\textbf{Speedup (vs xPU)}} &
\multirow{2}{*}{\shortstack{\textbf{EL/EG}\\\textbf{Speedup}}} \\
\cmidrule(lr){3-4}
\cmidrule(lr){5-6}
\cmidrule(lr){7-8}
& & \textbf{EG} & \textbf{EL}
  & \textbf{EG} & \textbf{EL}
  & \textbf{EG} & \textbf{EL} & \\
\midrule
1k  & Full   & 21 & 54 & 0.29 & 0.94 & 3.06$\times$ & 7.08$\times$ & 2.31$\times$ \\
2k  & Full   & 18 & 50 & 0.28 & 0.93 & 2.70$\times$ & 6.29$\times$ & 2.33$\times$ \\
4k  & Full   & 14 & 41 & 0.28 & 0.86 & 2.37$\times$ & 4.94$\times$ & 2.09$\times$ \\
\midrule
\multirow{2}{*}{8k}
    & Full   & 5  & 25 & 0.13 & 0.57 & 1.23$\times$ & 2.68$\times$ & 2.18$\times$ \\
    & Sparse & 14 & 41 & 0.18 & 0.57 & 1.54$\times$ & 2.79$\times$ & 1.81$\times$ \\
\midrule
\multirow{2}{*}{16k}
    & Full   & 0  & 0  & 0.11 & --   & 0.90$\times$ & --           & --           \\
    & Sparse & 14 & 41 & 0.14 & 0.46 & 1.09$\times$ & 1.87$\times$ & 1.72$\times$ \\
\bottomrule
\end{tabular*}
\vspace{-2mm}
\end{table}
}

\newcommand{\tabnine}{
\begin{table}[t]
\centering
\footnotesize
\setlength{\tabcolsep}{1.8pt}
\caption{Phase-wise power and energy efficiency (Qwen3-30B-A3B, mixed dataset, BS=4, 8GB HB).
Total includes 0.55\,W DRAM static and $\leq$0.02\,W on-chip comm. power.}
\label{tab:power_energy}
\begin{tabular*}{\linewidth}{@{\extracolsep{\fill}}llccccccc}
\toprule
\multirow{2}{*}{\textbf{Design}} &
\multirow{2}{*}{\textbf{Phase}} &
\multirow{2}{*}{\shortstack{\textbf{Lat.}\\(ms)}} &
\multicolumn{4}{c}{\textbf{Power (W)}} &
\multirow{2}{*}{\shortstack{\textbf{Thpt.}\\(tok/s)}} &
\multirow{2}{*}{\shortstack{\textbf{Energy}\\(mJ/tok)}} \\
\cmidrule(lr){4-7}
& & & \textbf{Comp.} & \textbf{HB} & \textbf{EXT} & \textbf{Total} & & \\
\midrule
xPU & AR & 70.2 & 0.01 & -- & 3.32 & 3.91 & 57.0 & 68.6 \\
\midrule
\multirow{3}{*}{ELMoE-3D}
  & Draft ($\times$10) & 33.1 & 0.95 & 5.64 & 0.00 & 7.15 &
    \multirow{3}{*}{271.6} & \multirow{3}{*}{14.7} \\
  & Verify & 115.4 & 0.22 & 0.74 & 1.55 & 3.08 & & \\
\cmidrule(lr){2-7}
  & \textit{Time-avg} & 148.5 & 0.38 & 1.83 & 1.21 & 3.99 & & \\
\bottomrule
\end{tabular*}
\vspace{-2mm}
\end{table}
}

%% file: Floats/algorithm.tex
\newcommand{\algone}{
\begin{algorithm}[t]
\footnotesize
\caption{HB-Aware Hybrid AR/SD Serving}
\label{alg:hb_serving}
\begin{algorithmic}[1]
\Require Batched tokens $\mathcal{B}$, Cached expert set $\mathcal{C}$
\Ensure Output tokens $\mathcal{Y}$
\Statex \textbf{HB (on-chip)}: DenseLayer, $\mathrm{MSB}(\mathcal{C})$
\Statex \textbf{EXT (off-chip)}: $\mathrm{LSB}(\mathcal{C})$, uncached experts
\Statex \textbf{Mode}: Offline-calibrated threshold $\tau$

\If{$|\mathcal{B}| \ge \tau$} \Comment{\textcolor{blue}{Speculative Decoding}}

\For{draft depth} \Comment{\textcolor{blue}{Draft Phase}}
    \State $x \gets \textsc{Embed}(\mathcal{B})$
    \For{layer $\ell$}
        \State $x \gets \textsc{DenseLayer}(x)$
        \State $\mathcal{G}_{\mathrm{thr}} \gets \textsc{TopK}_{\mathcal{C}}(\textsc{Route}(x))$ \Comment{Expert Throttling}
        \ForAll{$e \in \mathrm{Experts}[\mathcal{G}_{\mathrm{thr}}]$}
            \State $x \gets x + e(x;\; \mathrm{MSB}(\textsc{HB}[e]) \,\|\, \mathbf{1})$ \Comment{LSB Aug.}
        \EndFor
        \State $\textsc{RecordHotness}()$
    \EndFor
    \State $\mathcal{B} \gets \textsc{Sample}(x)$
\EndFor

\State $x \gets \textsc{Embed}(\mathcal{B})$ \Comment{\textcolor{blue}{Verify Phase}}
\For{layer $\ell$}
    \State $x \gets \textsc{DenseLayer}(x)$
    \State $\mathcal{G} \gets \textsc{TopK}\!\bigl(\textsc{Route}(x))$ \Comment{Original Routing}
    \ForAll{$e \in \mathrm{Experts}[\mathcal{G}]$}
        \If{$e \in \mathcal{C}$}
            \State $x \gets x + e(x;\; \mathrm{MSB}(\textsc{HB}[e]) \,\|\, \mathbf{0})$ \Comment{LSB Aug.}
            \State $x \gets x + e(x;\; \mathrm{LSB}(\textsc{EXT}[e]))$
        \Else
            \State $x \gets x + e(x;\; \textsc{EXT}[e])$
        \EndIf
    \EndFor
    \State $\textsc{UpdateCache}(\mathcal{C}_{\mathrm{next}})$
\EndFor
\State $\mathcal{Y} \gets \textsc{GreedyTreeVerify}(x,\mathcal{B})$

\Else \Comment{\textcolor{blue}{Autoregressive Decoding}}
\State $\mathcal{Y} \gets \textsc{Decode}(\mathcal{B})$ 
\EndIf

\State \Return $\mathcal{Y}$
\end{algorithmic}
\end{algorithm}
}